\pdfoutput=1
\PassOptionsToPackage{hyphens}{url}
\documentclass[11pt,letterpaper,logo]{preprint}

\usepackage[utf8]{inputenc}
\usepackage[T1]{fontenc}
\usepackage{microtype}
\usepackage{natbib}
\usepackage{latexsym}
\usepackage{graphicx}
\usepackage{xcolor}
\usepackage{pifont}
\usepackage{fontawesome5}
\newcommand{\buttonface}[2]{\tikz[baseline=-0.6ex]{\node[draw=AWSGray350, fill=white, rounded corners=8pt, line width=0.5pt, inner xsep=8pt, inner ysep=3pt, font=\papersans\color{PaperInk}] {#1\hspace{3pt} #2};}}
\newcommand{\linkbutton}[3]{\ifblank{#1}{\buttonface{#2}{#3}}{\href{#1}{\buttonface{#2}{#3}}}}

\hypersetup{hidelinks}

\usepackage{needspace}
\newtcolorbox{highlightbox}{enhanced, breakable, colback=AWSViolet50, colframe=PaperAccent,
  boxrule=0pt, leftrule=3pt, arc=6pt, left=7pt, right=7pt, top=6pt, bottom=6pt,
  before skip=6pt, after skip=6pt,
  before upper={\setlength{\parskip}{4pt}\setlength{\topsep}{0pt}}}
\newtcolorbox{defbox}{enhanced, breakable, colback=AWSViolet50, colframe=PaperAccent,
  boxrule=0pt, leftrule=3pt, arc=6pt, left=7pt, right=7pt, top=6pt, bottom=6pt,
  before skip=6pt, after skip=6pt,
  before upper={\setlength{\parskip}{4pt}\setlength{\topsep}{0pt}}}
\newcommand{\boxfont}{\fontsize{10.5}{12.98}\selectfont}
\tcbset{findingband/.style={enhanced, boxsep=0pt, left=3pt, right=4pt, top=1.6pt, bottom=1.6pt, arc=3pt,
  colback=AWSViolet50, colframe=PaperAccent, boxrule=0pt, leftrule=2pt, width=\linewidth,
  before skip=\dimexpr\parskip+5pt\relax, after skip=0pt, after={\par\nobreak\vskip\dimexpr 2.5pt-\parskip\relax\noindent}}}
\newcommand{\finding}[1]{\begin{tcolorbox}[findingband]\textbf{#1}\end{tcolorbox}\ignorespaces}
\newcommand{\pullup}{\begingroup\hyphenpenalty=20\exhyphenpenalty=20\spaceskip=\fontdimen2\font plus \fontdimen3\font minus 1.5\fontdimen4\font\looseness=-1\relax}
\newcommand{\stoppullup}{\par\endgroup}
\newcommand{\findingw}[1]{\begin{tcolorbox}[findingband,after={\par\nobreak\vskip\dimexpr 2.5pt-\parskip\relax}]\textbf{#1}\end{tcolorbox}}

\AtBeginDocument{\tcbset{pbox/.append style={before skip=10pt, after skip=10pt}}}

\newcommand{\myTool}{\textsc{Inventa}}
\definecolor{exGray}{HTML}{6B7280}

\newcommand{\rnd}[1]{\smash[t]{#1}}
\usepackage{booktabs}
\usepackage{colortbl}
\usepackage{multirow}
\newcommand{\best}[1]{\textbf{#1}}
\newcommand{\nd}[1]{\underline{#1}}

\definecolor{tbGroup}{HTML}{E3E6EA}\definecolor{tbSub}{HTML}{F3F4F6}\definecolor{tbOurs}{HTML}{E3EEFB}\definecolor{tbHead}{HTML}{F4F6F8}
\newcommand{\oursL}[1]{\multicolumn{1}{@{}>{\columncolor{tbOurs}[0pt][\tabcolsep]}l!{\color{black!22}\vrule}}{#1}}\newcommand{\oursR}[1]{\multicolumn{1}{>{\columncolor{tbOurs}[\tabcolsep][0pt]\numsz}c@{}}{#1}}

\newcommand{\bandstrut}{\rule[-0.7ex]{0pt}{2.4ex}}\newcommand{\substrut}{\rule[-0.55ex]{0pt}{2.15ex}}
\newcommand{\grp}[2]{\multicolumn{#1}{@{}>{\columncolor{tbGroup}[0pt][\tabcolsep]}l}{\bandstrut\textbf{#2}}}
\newcommand{\grpours}[2]{\multicolumn{#1}{@{}>{\columncolor{tbOurs}[0pt][\tabcolsep]}l}{\bandstrut\textbf{#2}}}
\newcommand{\ind}{\hspace{0.6em}}
\newcommand{\mh}[1]{{\scriptsize\bfseries #1}}
\newcommand{\numsz}{\fontsize{11}{11}\selectfont}\newcolumntype{N}{>{\numsz}c}\newcolumntype{R}{>{\numsz}r}
\newcommand{\mhh}[2]{\mh{\begin{tabular}[b]{@{}c@{}}#1\\#2\end{tabular}}}
\usepackage{tikz}
\usepackage{amsmath,amssymb}
\usepackage{amsthm}
\newtheoremstyle{defnamed}{\topsep}{\topsep}{}{}{\bfseries}{.}{.5em}%
  {\thmname{#1}\thmnumber{ #2}\thmnote{ (\bfseries\itshape #3\normalfont)}}
\theoremstyle{defnamed}\newtheorem{definition}{Definition}[section]
\usepackage{enumitem}
\newlist{stages}{enumerate}{1}
\setlist[stages]{label=(\roman*),wide=0pt,leftmargin=1em,nosep,topsep=1pt}
\newlist{actions}{itemize}{1}
\setlist[actions]{label=\textbullet,wide=0pt,leftmargin=1em,nosep,topsep=1pt}
\usepackage{algorithm}
\makeatletter\g@addto@macro\fs@ruled{\def\@fs@post{\kern2pt\hrule\relax}}\makeatother
\usepackage[italicComments=false,commentColor=black!55]{algpseudocodex}
\algrenewcommand\algorithmicloop{\textbf{atomic}}
\makeatletter\renewcommand{\algpx@commentFormat}[1]{{\fontsize{7.6}{8}\selectfont\itshape\textcolor{\algpx@commentColor}{#1}}}\expandafter\patchcmd\csname\string\algorithmic\endcsname{\labelwidth 1.2em}{\labelwidth 1.5em}{}{\typeout{ALGPATCHFAIL}}\makeatother
\tikzset{algpxIndentLine/.style={draw=black!22,line width=0.4pt}}
\DeclareCaptionFont{algcap}{\fontsize{9}{10.6}\selectfont}
\floatstyle{plaintop}\restylefloat{algorithm}
\newtcolorbox{algobox}{enhanced, colback=AWSViolet50, colframe=PaperAccent, boxrule=0pt, leftrule=3pt, arc=6pt, left=7pt, right=7pt, top=4pt, bottom=3.5pt, before skip=0pt, after skip=0pt}
\newlength{\algW}\newcommand{\algkw}[1]{\textcolor{PaperAccent}{\textbf{#1}}}
\algrenewcommand\algorithmicrequire{\algkw{Input:}}\algrenewcommand\algorithmicensure{\algkw{Output:}}
\usetikzlibrary{calc}
\usepackage{tcolorbox}\tcbuselibrary{skins,breakable}
\definecolor{promptO}{HTML}{2E4F7F}
\tcbset{pbox/.style={enhanced, breakable, colback=white, colframe=#1, skin first=enhanced, skin middle=enhanced, skin last=enhanced, colbacktitle=#1, coltitle=white,
  boxrule=0.55pt, arc=1.2mm, boxsep=0pt, left=1.6mm, right=1.6mm, top=1.1mm, bottom=1.2mm, toptitle=0.9mm, bottomtitle=0.8mm,
  fonttitle=\fontsize{7.4}{8.6}\selectfont\bfseries, fontupper=\fontsize{6.9}{8.25}\selectfont,
  before upper={\setlength{\parskip}{1.8pt}\setlength{\parindent}{0pt}\emergencystretch=1.5em},
  before skip=4pt, after skip=6pt}}
\newcommand{\fn}[1]{{\upshape\fontfamily{zi4}\fontsize{8.2}{10}\selectfont #1}}
\newcommand{\ph}[1]{{\color{ridU!75!black}$\langle$\textit{#1}$\rangle$}}
\newcommand{\mdh}[1]{{\color{black!45}\fn{\#\#}}~\textbf{#1}}
\newcommand{\pnote}[1]{{\itshape\color{black!72}#1}\par}
\newcommand{\plab}[2]{\par\vspace{2.4pt}\noindent{\color{#1}\fontsize{6.1}{7}\selectfont\sffamily\bfseries #2}\hspace{3pt}{\color{#1!30}\leaders\vrule height 2.4pt depth -1.95pt\hfill\kern0pt}\par\nobreak\vspace{0.4pt}\nobreak}
\usepackage{fancyvrb}
\newtcolorbox{pcodebox}{enhanced, colback=black!5, frame hidden, arc=0pt, boxrule=0pt, boxsep=0pt, left=2pt, right=2pt, top=1.6pt, bottom=1.6pt, before skip=2pt, after skip=2pt}
\newcommand{\pk}[1]{{\color{promptO}\bfseries #1}}
\newcommand{\pc}[1]{{\color{black!50}#1}}
\newcommand{\pcode}[1]{\par{\setlength{\fboxsep}{2pt}\colorbox{black!5}{\parbox{\dimexpr\linewidth-4pt\relax}{\raggedright #1}}}\par}
\SetEnumitemKey{pfiles}{nosep, style=unboxed, leftmargin=0.9em, labelsep=0.45em, before={\endgraf\raggedright}, font=\normalfont\fontfamily{zi4}\fontsize{8.2}{10}\selectfont}
\SetEnumitemKey{pacts}{nosep, style=unboxed, leftmargin=0.9em, labelsep=0.45em, before={\endgraf\raggedright}, font=\normalfont\bfseries\scshape\color{promptO}}
\definecolor{bgU}{HTML}{93B9EC}\definecolor{bgV}{HTML}{BCDD86}\definecolor{bgK}{HTML}{F3B866}
\newcommand{\lgsq}[1]{\textcolor{#1}{\rule[-0.1ex]{1.3ex}{1.3ex}}\,}\newcommand{\lgU}[1]{\lgsq{bgU}#1}\newcommand{\lgV}[1]{\lgsq{bgV}#1}\newcommand{\lgK}[1]{\lgsq{bgK}#1}
\newcommand{\amark}[1]{\tikz[remember picture,overlay]\coordinate(#1);}
\newcommand{\ashade}[4][0]{\begin{tikzpicture}[remember picture,overlay]\fill[#2,blend mode=multiply]($(algL|-#3)+(-1.5pt,0.74\baselineskip)$)rectangle($(algL|-#4)+(14.2pt,-0.26\baselineskip-#1\baselineskip)$);\end{tikzpicture}}
\newcommand{\Cert}{\textsc{Cert}}
\newcommand{\Score}{\textsc{Score}}
\newcommand{\Cplx}{\textsc{Cplx}}
\newcommand{\Gain}{\textsc{Gain}}
\newcommand{\Valid}{\textsc{Valid}}

\newcommand{\Drop}{\textsc{Abl}}
\newcommand{\DropM}{\textsc{Abl}^{\max}}
\newcommand{\Ood}{\textsc{OOD}}
\newcommand{\Viol}{\mathcal{C}_{\mathrm{viol}}}
\newcommand{\Undec}{\mathcal{C}_{\mathrm{undec}}}

\newcommand{\rid}[1]{\tikz[baseline=(c.base)]\node[circle,draw=black!65,line width=0.4pt,inner sep=0pt,minimum size=6.4pt,font=\fontsize{\ridfs}{\ridfs}\selectfont\upshape](c){#1};\,}\newcommand{\ridfs}{4.6}
\definecolor{ridU}{HTML}{3F70B8}\definecolor{ridV}{HTML}{5F9128}\definecolor{ridK}{HTML}{D28418}
\newcommand{\ridc}[2]{\tikz[baseline=(c.base)]\node[circle,fill=#2,inner sep=0pt,minimum size=7.5pt,text=white,font=\fontsize{\ridfs}{\ridfs}\selectfont\sffamily\bfseries\upshape](c){#1};\,}
\newcommand{\subgrp}[2]{\multicolumn{#1}{@{}>{\columncolor{tbSub}[0pt][\tabcolsep]}l}{\substrut\ind\textit{#2}}}

\usepackage{float}
\usepackage{wrapfig}
\newcommand{\asc}{945}\newcommand{\wJ}{0.36\textwidth}\newcommand{\wA}{0.62\textwidth}\newcommand{\wW}{0.535\textwidth}\newcommand{\wR}{0.445\textwidth}\newcommand{\gapC}{6pt}\newcommand{\rsJ}{1.09}
\newcommand{\tsc}{685}\newcommand{\wEB}{0.4967\textwidth}\newcommand{\wSLB}{0.4813\textwidth}\newcommand{\wSC}{0.462\textwidth}
\newcommand{\fitat}[3]{\begingroup\setlength{\tabcolsep}{0pt}\sbox0{#3}\setlength{\tabcolsep}{\dimexpr(\linewidth*1000/#1-\wd0)/#2\relax}\scalebox{0.#1}{#3}\endgroup}
\newcommand{\rsEB}{1.1405}\newcommand{\rsSLB}{1}\newcommand{\gapIS}{6pt}\newcommand{\wSCtop}{-6pt}\newcommand{\wSCbot}{-3pt}\newcommand{\rsSC}{1.07}
\AtBeginEnvironment{table}{\setlength{\extrarowheight}{1.5pt}}\AtBeginEnvironment{table*}{\setlength{\extrarowheight}{1.5pt}}
\newsavebox\ndbox\renewcommand{\nd}[1]{\leavevmode\sbox\ndbox{#1}\rlap{\rule[-0.9pt]{\wd\ndbox}{0.4pt}}\usebox\ndbox}
\tcbset{pbox/.append style={fonttitle=\fontsize{9}{11}\selectfont\bfseries,
  fontupper=\fontsize{8.6}{10.6}\selectfont,
  before upper={\setlength{\parskip}{3pt}\setlength{\parindent}{0pt}\emergencystretch=1.5em},
  left=2.5mm, right=2.5mm, top=1.8mm, bottom=1.8mm}}
\renewcommand{\plab}[2]{\par\vspace{3.5pt}\noindent{\color{#1}\fontsize{7.6}{9}\selectfont\sffamily\bfseries #2}\hspace{3pt}{\color{#1!30}\leaders\vrule height 3pt depth -2.5pt\hfill\kern0pt}\par\nobreak\vspace{0.6pt}\nobreak}

\colorlet{PaperTitleBack}{AWSGray200}

\title{\myTool{}: Decentralized Self-Improving\\
Agents for Autonomous Scientific Discovery}

\runningtitle{Inventa: Decentralized Self-Improving Agents for Autonomous Scientific Discovery}

\paperlinks{\linkbutton{https://jprithwish.github.io/inventa/}{\textcolor{PaperAccent}{\faGlobe}}{Project Page}}

\paperlogos{\includegraphics[height=22pt]{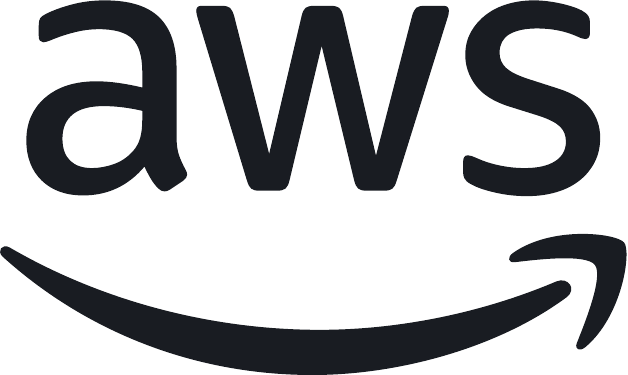}}

\hypersetup{
  pdftitle={
    Inventa: Decentralized Self-Improving Agents for Autonomous Scientific Discovery
  },
  pdfauthor={
    Prithwish Jana,
    Nikos Kanakaris,
    Purak Jain,
    Anoop Deoras,
    Sahika Genc
  },
  pdfsubject={Artificial Intelligence},
  pdfkeywords={
    automated scientific discovery,
    research ideation,
    evolutionary search,
    multi-agent systems,
    language model agents
  }
}

\begin{document}

\author{
  \normalfont
  \begin{minipage}{0.98\linewidth}
    \raggedright
    \setlength{\parskip}{0pt}

    \textbf{Prithwish~Jana}$^{1,\dagger}$,
    \textbf{Nikos~Kanakaris}$^{2}$,
    \textbf{Purak~Jain}$^{2}$,
    \textbf{Anoop~Deoras}$^{2}$,
    \textbf{Sahika~Genc}$^{2}$

    \par\vspace{7pt}

    \mbox{$^{1}$Georgia Institute of Technology}
    \quad
    \mbox{$^{2}$Amazon}

  \end{minipage}
}

\begin{abstract}
\textbf{Abstract:} \looseness=-1 Large language model (LLM) agents can generate research ideas and improve scientific discovery through iterative experimentation. However, increasing the number of agents or iterations does not automatically translate to producing more diverse or feasible ideas. Agents may converge on similar candidates, propagate unsupported claims through shared memory components and repeatedly apply search strategies that no longer yield any improvements.

We introduce \myTool{}, a framework for automated scientific discovery that employs autonomous research units and verified shared knowledge. Each unit creates a lineage of candidate ideas using a mutator agent. A local orchestrator agent periodically adapts the mutator's harness. When progress stalls, the orchestrator agent replaces its model or restarts the idea evolution process. Verifiers evaluate each candidate idea and record verified improvements, useful components and constraint violations in a shared knowledge, the only channel between units. We evaluate \myTool{} on five benchmarks spanning the research cycle, covering retrieval (\mbox{ScholarCatalyst}), ideation (IdeaBench), proposal writing (\mbox{IdeaScientist}), and scientific law discovery (\mbox{EurekaBench}, SciLaws-Bench). It leads all five, achieving 73.6\% on EurekaBench versus 56.2\% for CORAL. On IdeaBench, it discovers 26.0 novel ideas per paper, compared with 18.6 for ResearchAgent, with 82\% fewer model calls.
\end{abstract}

\maketitle

{\let\thefootnote\relax
\footnotetext{%
  $^{\dagger}$Work done while at Amazon.
}}

\section{Introduction}
\label{sec:intro}
\begin{figure*}[t]
  \centering
  \includegraphics[width=\textwidth]{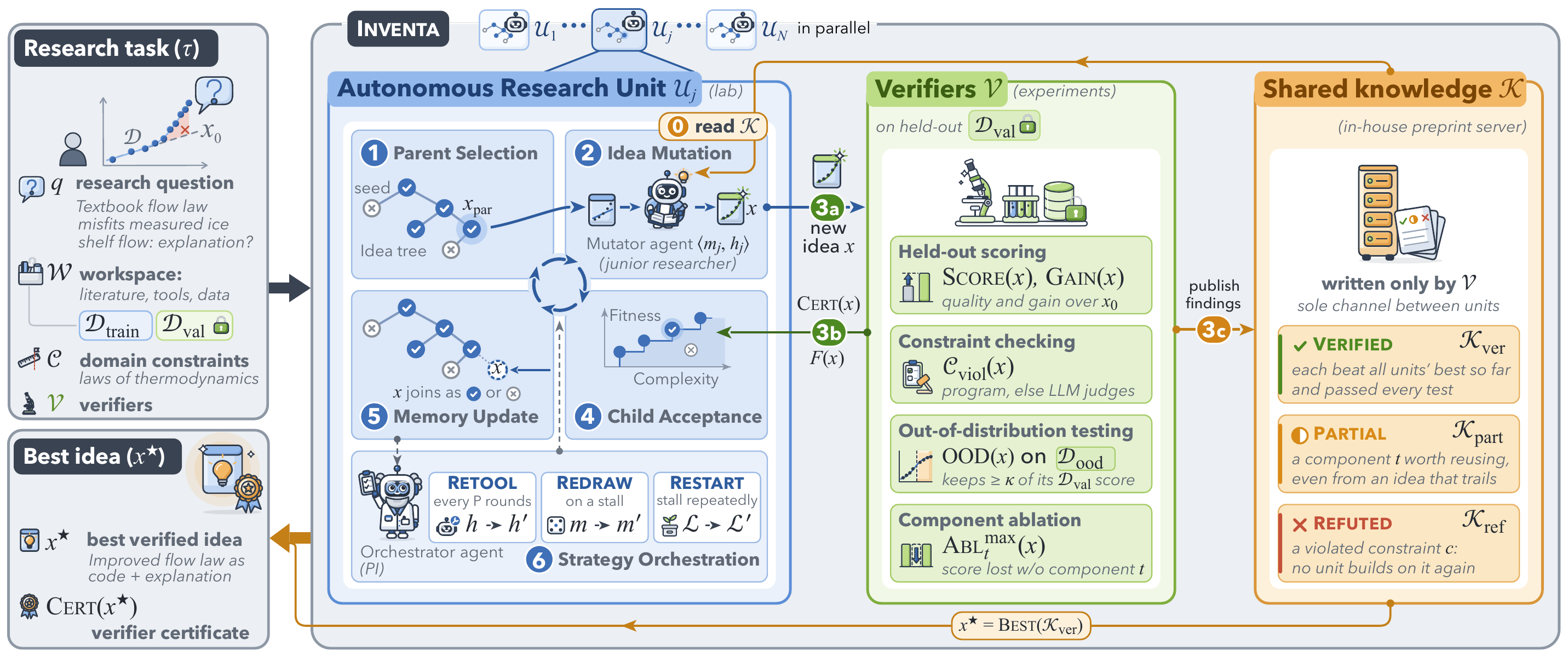}
  \vspace{-7.5mm}
  \caption{\textbf{Overview of \myTool{}.} Given a research task, \myTool{} generates novel and verified ideas. Decentralized \textit{autonomous research units} explore candidate ideas in parallel. Each unit \protect\ridc{0}{ridK}\unkern{}~reads the \textit{shared knowledge} ($\mathcal{K}$), \protect\ridc{1}{ridU}\unkern{}~selects a parent from its idea tree, and \protect\ridc{2}{ridU}\unkern{}~uses a mutator agent to propose a child idea. \textit{Verifiers} \protect\ridc{3a}{ridV}\unkern{}~evaluate it with held-out scoring, constraint checks, and out-of-distribution or ablation tests, \protect\ridc{3b}{ridV}\unkern{}~return its certificate, and \protect\ridc{3c}{ridK}\unkern{}~publish qualifying findings to $\mathcal{K}$. The unit \protect\ridc{4}{ridU}\unkern{}~applies acceptance checks and \protect\ridc{5}{ridU}\unkern{}~records the child in its idea tree. A per-unit orchestrator \protect\ridc{6}{ridU}\unkern{}~evolves the mutator regularly. $\mathcal{K}$ is the sole channel for inter-unit communication, and only verifiers write to it.}
  \label{fig:method}
\end{figure*}

Large language models (LLMs) are increasingly used to assist in scientific research, ranging from identifying relevant literature to proposing candidate hypotheses and implementing experiments. Evolutionary search extends these capabilities by using evaluation feedback to refine candidates. Systems such as AlphaEvolve~\citep{novikov2025alphaevolve}, EvoX~\citep{liu2026evox}, CORAL~\citep{qu2026coral}, and MILO~\citep{jana2026milo} show the potential of combining LLMs with closed-loop search. These advances motivate automated research assistants that can explore novel ideas, evaluate their feasibility and retain useful findings across experiments.

\begingroup\spaceskip=\fontdimen2\font plus \fontdimen3\font minus 1.2\fontdimen4=-1\relax\textbf{Automated research assistance} involves several related tasks across the complete scientific research cycle. These include \emph{retrieving prior work} that provides inspiration for a research question~\citep{kim2026scholarcatalyst}, \emph{generating ideas} from existing papers~\citep{guo2024ideabench}, \emph{assessing their novelty} against related work~\citep{schopf2026rinobench}, and \emph{discovering scientific laws} from data~\citep{huang2026scilaws}. They differ in output but share a common challenge: generating promising candidates is insufficient unless a system can also verify them and use the evidence to guide further exploration.\stoppullup

Recent \textit{AI co-scientists}, \looseness=-1 such as the Agentic Scientist~\citep{lange2026agentic} and InternAgent-1.5~\citep{feng2026internagent}, coordinate multiple agents to generate, test, and refine hypotheses. Despite this progress, several challenges remain unresolved. \textbf{First}, \textbf{\textit{maintaining diversity as search grows}}. These systems largely follow myopic hill-climbing to optimize the most promising candidates, which narrows the exploration radius~\citep{lange2026agentic}. Furthermore, additional samples or iterations yield increasingly similar candidates, so extra computation brings diminishing returns~\citep{jiang2025artificial,si2025can}. \textbf{Second}, \textbf{\textit{sharing useful knowledge without propagating errors}}. Isolated agents cannot benefit from one another's findings, whereas interacting agents often converge on similar ideas~\citep{chen2026diversity}. When shared information is not independently verified, unsupported claims also influence subsequent attempts~\citep{kim2025towards}. \textbf{Third}, \textbf{\textit{adapting the proposal mechanism}}. Refining candidate ideas does not, by itself, change the prompts, tools, or control flow used to generate them, so iterations can improve ideas but not expand the search space. \textbf{Fourth}, \textbf{\textit{satisfying scientific constraints}}. Highly novel ideas often prove infeasible under the task's assumptions or physical requirements, creating a novelty--feasibility gap~\citep{geng2026eurekabench}.

To address these challenges, we propose \myTool{} (Fig.~\ref{fig:method}), a framework that combines i)~autonomous research units, ii)~self-adaptive proposal mechanisms and iii)~sharing of verified findings. We draw inspiration from how science advances in practice: research groups or labs pursue different approaches in a decentralized fashion while building on evidence reported by others. Each unit (like a lab) maintains its own idea lineage and uses a mutator agent (junior researcher) to develop new candidates. A per-unit orchestrator (the lab's principal investigator, or PI) adapts the mutator's harness and intervenes when progress stalls. Verifiers evaluate the candidates, return structured feedback, and add verified findings to the shared knowledge.

The proposed framework separates three operations: developing candidate ideas, self-adapting the process that develops them, and determining which findings can be shared. This separation allows units to pursue different approaches while receiving, only through shared knowledge, common evidence about successful candidates, useful components, and observed constraint violations.

\Needspace*{14\baselineskip}
\begin{highlightbox}\boxfont
\noindent{\papersans\color{PaperInk} Contributions.} Our main contributions are:\par\noindent\mbox{\scalebox{1.2}{\ding{182}}~\textbf{Self-improving}}\textbf{ idea generation.} We propose \myTool{}, an agentic framework for scaling scientific discovery and research ideation. The framework refines ideas while continually evolving the agents that generate them, expanding the space of ideas it can reach as the search proceeds.\par\noindent\mbox{\scalebox{1.2}{\ding{183}}~\textbf{Diverse}}\textbf{ exploration with verified knowledge sharing.} We introduce autonomous research units that explore different approaches, each with its own history of ideas and experiments. Units communicate only through shared knowledge written by verifiers: positive findings point to promising directions, while refuted ones, such as constraint violations, prune the search space.\par\noindent\mbox{\scalebox{1.2}{\ding{184}}~\textbf{Extensive}}\textbf{ \mbox{experiments.}} On five benchmarks from retrieving prior work to scientific law discovery, \myTool{} leads every primary metric by 6--40\% and alone passes every EurekaBench constraint test (Fig.~\ref{fig:cycle}).
\end{highlightbox}

\begin{figure*}[t]
  \centering
  \definecolor{figslate}{HTML}{2F3742}
  \begin{tikzpicture}[x=1mm,y=1mm]
    \node[anchor=north west,inner sep=0] at (0,0) {\includegraphics[width=\textwidth]{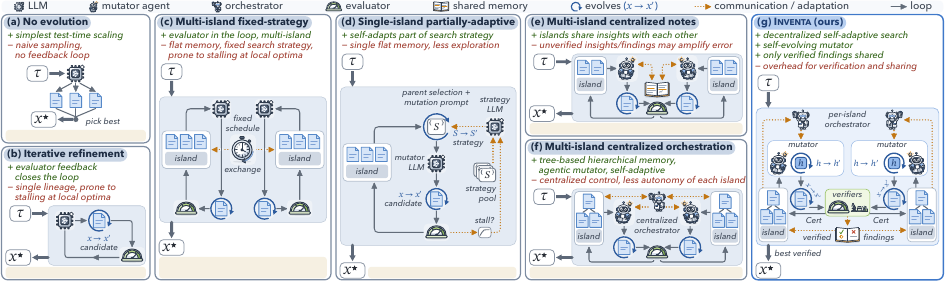}};
    \tikzset{nm/.style={anchor=center,inner sep=0,align=center,
      font=\fontfamily{phv}\fontsize{4}{4.8}\selectfont,text=figslate}}
    \hypersetup{citecolor=figslate,linkcolor=figslate}%
    \node[nm,text width=23.7mm] at (12.8,-23.0) {Best-of-\textit{N}~\citeyearpar{brown2024large}};
    \node[nm,text width=23.7mm] at (12.8,-46.2) {ResearchAgent~\citeyearpar{baek2025researchagent}, AIDE2~\citeyearpar{srikanth2026aide2}};
    \node[nm,text width=28.4mm] at (41.1,-45.3) {FunSearch~\citeyearpar{romeraparedes2024mathematical},~~AlphaEvolve~\citeyearpar{novikov2025alphaevolve},\\ OpenEvolve~\citeyearpar{sharma2025openevolve}};
    \node[nm,text width=25.1mm] at (74.9,-46.2) {EvoX~(\hyperlink{cite.liu2026evox}{2026})};
    \node[nm,text width=36.1mm] at (107.7,-21.8) {AgentRxiv~\citeyearpar{schmidgall2025agentrxiv}, CORAL~\citeyearpar{qu2026coral}};
    \node[nm,text width=36.1mm] at (107.7,-46.2) {MILO~\citeyearpar{jana2026milo}};
  \end{tikzpicture}
  \vspace{-7mm}
  \caption{\textbf{Comparison of search mechanisms.} Prior methods (a)--(f) range from repeated sampling to island-based evolution with fixed, partially adaptive, or centralized strategies; all keep the mutator fixed. In \myTool{} (g), each decentralized unit has a local orchestrator that evolves its mutator, while~only~\mbox{verifiers}~publish~shared~findings.}
  \label{fig:relwork}
\end{figure*}

\section{Related Work}
\label{sec:related}

\noindent\textbf{Scaling search.} AI scientists generate and test ideas~\citep{lu2026towards,gottweis2026accelerating}, often by LLM-guided search. Yet in equation discovery, evolution matches fresh sampling at equal budgets~\citep{li2026dictionaries}, and repeated runs or added islands stall in idea basins~\citep{cai2026fork,antoniades2026heuresis} or collapse onto a few principles~\citep{ran2026principle}. \myTool{} runs many units that redraw their model or restart their lineage on a stall.

\noindent\textbf{Adapting the generator.} \looseness=-1 AlphaEvolve co-evolves its prompts, CORAL's agents write their own skills, and MILO and ShinkaEvolve pick among fixed agents or LLMs~\citep{novikov2025alphaevolve,qu2026coral,jana2026milo,lange2025shinkaevolve} (Fig.~\ref{fig:relwork}). Harness search optimizes a final harness on a fixed search set~\citep{lee2026metaharness,zhang2025darwin}, where overfitting explains about half the gain~\citep{ding2026harness}. In \myTool{}, an orchestrator acts as a meta-mutator, evolving the mutator that evolves ideas during search and keeping an edit only if twin trials show significantly fitter children.

\noindent\textbf{Sharing verified knowledge.} In CORAL and AgentRxiv, agents write shared state~\citep{qu2026coral,schmidgall2025agentrxiv}; such state can spread reward hacks~\citep{hu2026reward} and early consensus~\citep{yoon2026arcticswarm}. DuoDiscover's agents optimize self-built proxy evaluators~\citep{agarwal2026duodiscover}; in \myTool{}, only verifiers write shared knowledge, scoring on data no mutator sees, and publish refuted findings that prune all units' search.

\section{Preliminaries and Problem Statement}
\label{sec:prelim}

\begin{defbox}\boxfont
\begin{definition}[Agents and harnesses]
\label{def:agents}
An AI agent $A:=\langle m,h\rangle$ consists of an LLM $m$ and a harness $h$. The harness specifies the agent's prompts, tools, context management, and control flow. In our setting, we keep the model weights frozen and allow the harness to change during evolution. Let $\mathcal{M}$ denote the set of available models and $\mathcal{H}$ the initial set of harnesses. The pool $\mathcal{G}\subseteq\mathcal{M}\times\mathcal{H}$ contains their compatible model--harness pairs.
\end{definition}
\end{defbox}

\begin{defbox}\boxfont
\begin{definition}[Research tasks and candidate ideas]
\label{def:tasks}
A \textbf{research task}, the input to \myTool{}, is a tuple $\tau:=\langle q,\mathcal{W},\mathcal{C},\mathcal{V}\rangle$, where $q$ is a research question; $\mathcal{W}$ is a workspace of literature, tools, and data $\mathcal{D}$; $\mathcal{C}$ is a set of task-specific constraints; and $\mathcal{V}$ is a set of verifiers that run a full experiment on each candidate, executing it on held-out data or in task simulators, testing task constraints, and using LLM judges where programmatic tests are unavailable. Given $\tau$, \myTool{} outputs a \textbf{candidate idea} $x$ that answers $q$ in the task's format, such as code with an explanation; $T(x)$ denotes its components, such as the terms of a law, which verifiers ablate (\S\ref{sec:verifiers}). We show an example task in Fig.~\ref{fig:method} (left) and examples of all task classes in Appendix~\ref{app:examples}.
\end{definition}
\end{defbox}

\begin{defbox}\boxfont
\begin{definition}[Data splits and candidate evaluation]
\label{def:splits}
Research tasks distinguish \textit{data regimes} $\mathcal{D}_\rho\subseteq\mathcal{D}$. For out-of-distribution evaluation, one regime is held out as $\mathcal{D}_{\mathrm{ood}}$, and the rest form $\mathcal{D}_{\mathrm{id}}$. Agents fit candidates on the training set $\mathcal{D}_{\mathrm{train}}$ from $\mathcal{D}_{\mathrm{id}}$; only verifiers access the validation set $\mathcal{D}_{\mathrm{val}}$, which includes $\mathcal{D}_{\mathrm{ood}}$. A sealed test set $\mathcal{D}_{\mathrm{test}}$ scores only the final candidate and never guides search. During search, verifiers compute the task-specific metric $\Score(x):=\Score(x;\mathcal{D}_{\mathrm{val}})$, the improvement $\Gain(x):=\Score(x)-\Score(x_0)$ over a baseline $x_0$, and the fraction $\Valid(x)\in[0,1]$ of constraints without a confirmed violation.
\end{definition}
\end{defbox}

\begin{highlightbox}\boxfont
\noindent\textbf{Problem statement.}
Given a task $\tau$ and a search budget, \myTool{} seeks a candidate $x^\star$ that maximizes the validation score $\Score(x^\star)$ subject to $\Valid(x^\star)=1$ and $\Gain(x^\star)>\delta$ for a noise margin $\delta$, aiming to score high on the unseen $\mathcal{D}_{\mathrm{test}}$.
\end{highlightbox}

\section{Methodology}
\label{sec:method}

\begin{algorithm}[t]
\caption{\textbf{(\myTool{}).} $N$ \lgU{autonomous research units} evolve their mutators and share \lgV{verifier}-written \lgK{findings}.}
\label{alg:inventa}
\algrenewcommand\algorithmicindent{1em}%
\begin{algobox}
\fontsize{9}{10.4}\selectfont\renewcommand{\ridfs}{5.8}%
\setlength{\algW}{\linewidth}\noindent\amark{algL}\par\vspace*{-\dimexpr\baselineskip-2pt\relax}%
\begin{algorithmic}[1]
\Require \begin{tabular}[t]{@{}l@{}}task $\tau$ with verifiers $\mathcal{V}$, pool $\mathcal{G}$, budget $B$\\\textcolor{black!55}{\textit{hyperparameters:}} number of units $N$, patience $P$, lag $\lambda$\end{tabular}
\Ensure best verified idea $x^{\star}$ with $\Cert(x^{\star})$
\State \amark{k0}\rule{0pt}{\dimexpr0.74\baselineskip+2.8pt\relax}$\mathcal{K}=(\mathcal{K}_{\mathrm{ver}},\mathcal{K}_{\mathrm{part}},\mathcal{K}_{\mathrm{ref}})\gets(\emptyset,\emptyset,\emptyset)$ \Comment{shared knowledge}
\ForAll{$j=1,\dots,N$ \textbf{in parallel}} \Comment{autonomous units $\mathcal{U}_j$}
  \State \amark{u0}$\mu_j\sim\mathcal{G}$;\ $\mathcal{L}_j\gets\{\text{seed idea}\}$ \Comment{initialize $\mathcal{U}_j$}
  \For{$r_j\gets1$ \textbf{to} $B$}
    \State \textbf{wait until} $r_j<\min_{j'\in\mathrm{active}}r_{j'}+\lambda$ \Comment{bounded lag}
    \State \amark{k1}$\mathcal{K}_j\gets\textsc{Snapshot}(\mathcal{K})$ \Comment{\rid{0} read the shared knowledge}
    \State \amark{a1}$x_{\mathrm{par}}\gets\textsc{SelectParent}(\mathcal{L}_j)$ \Comment{\rid{1} select a parent}
    \State \amark{a2}$x\gets\mu_j(x_{\mathrm{par}},\Cert(x_{\mathrm{par}}),\mathcal{L}_j,\mathcal{K}_j)$ \Comment{\rid{2} write a child}
    \State \amark{b1}$\Cert(x),F(x)\gets\mathcal{V}(x)$ \Comment{\rid{3a} test, \rid{3b} return the certificate}
    \Loop\amark{c1} \Comment{\rid{3c} publish the findings}
      \State $\smash[t]{x^{\mathrm{best}}}\gets\textsc{Best}(\mathcal{K}_{\mathrm{ver}})$ \Comment{re-check the best verified idea}
      \State \amark{c2}$\mathcal{K}\gets\mathcal{K}\cup\textsc{Findings}(x,\smash[t]{x^{\mathrm{best}}})$ \Comment{share with all units}
    \EndLoop
    \State \amark{d1}$\ell_x\gets\textsc{Accept}(x,\mathcal{L}_j)$ \Comment{\rid{4} accept or reject the child}
    \State $\mathcal{L}_j\gets\mathcal{L}_j\cup\{(x,\ell_x)\}$ \Comment{\rid{5} record in the unit's memory}
    \If{$\mathcal{U}_j$ stalls} \Comment{no new best fitness in $P$ scored rounds}
      \State \textsc{Redraw} or \textsc{Restart} \Comment{\rid{6} change model or approach}
    \EndIf
    \State \amark{d2}\textbf{every} $P$ rounds: \textsc{Retool} \Comment{\rid{6} evolve harness via twin trial}
  \EndFor
\EndFor
\State \amark{k2}\Return $x^{\star}\gets\textsc{Best}(\mathcal{K}_{\mathrm{ver}})$ with $\Cert(x^{\star})$%
\begin{tikzpicture}[remember picture,overlay]\draw[black!28,line width=0.35pt]($(algL|-k0)+(-1.5pt,0.74\baselineskip+1.5pt)$)--++(\algW+1.5pt,0);\end{tikzpicture}\ashade{bgK}{k0}{k0}\ashade{bgU}{u0}{u0}\ashade{bgK}{k1}{k1}\ashade{bgU}{a1}{a2}\ashade{bgV}{b1}{b1}\ashade{bgK}{c1}{c2}\ashade{bgU}{d1}{d2}\ashade{bgK}{k2}{k2}
\end{algorithmic}
\end{algobox}
\vspace{-11pt}
\end{algorithm}
\myTool{} comprises autonomous research units, verifiers, and a shared knowledge (\S~\ref{sec:units}--\ref{sec:knowledge}). We detail the asynchronous evolution loop in Alg.~\ref{alg:inventa}.

\subsection{Autonomous Research Units}
\label{sec:units}

Each unit $\mathcal{U}_j$, where $j\in\{1,\dots,N\}$, \looseness=-1 advances at its own pace. Its \textit{round} $r_j$ counts the attempts it has made so far. At round $r_j=r$, we define its state as $\rnd{\mathcal{U}_j^{(r)}:=\langle\mathcal{L}_j^{(r)},\mu_j^{(r)}\rangle}$, consisting of its lineage memory and current mutator. Its orchestrator $\mathcal{O}_j$ adapts both the mutator and the unit's search direction.

\noindent\textbf{Lineage memory.}\label{sec:lineage}
The memory $\rnd{\mathcal{L}_j^{(r)}}$ \looseness=-1 is an \textit{idea tree} of the candidates $\mathcal{U}_j$ explored. Each \textit{node} holds an idea, its certificate of verifier evidence (\S\ref{sec:verifiers}), and an acceptance label. Each \textit{edge} links a parent to its child. The tree grows from a \textit{seed idea} proposed by $\mathcal{O}_j$; every later round $r_j=r$ has five stages:

\begin{stages}[label=\textbf{(\roman*)}]
\item \textbf{Parent selection.} To pick a parent idea $x_{\mathrm{par}}$, the unit treats the accepted nodes $\mathcal{A}$ of $\rnd{\mathcal{L}_j^{(r-1)}}$ as arms of a multi-armed bandit rewarded by the fitness $\rnd{F(x):=\Score(x)\times\mathbb{I}[\Valid(x)=1]}$. A node's value $\rnd{\bar s_v:=\tfrac{1}{n_v}\sum_{u\in\{v\}\cup\mathrm{ch}(v)}F(u)}$ averages the fitness of $v$ and its children $\mathrm{ch}(v)$, with $n_v:=1+|\mathrm{ch}(v)|$. By Thompson sampling~\citep{russo2018tutorial}, the unit draws ${\tilde s_v\sim\mathcal{N}(\bar s_v,\sigma^2/n_v)}$ and selects $x_{\mathrm{par}}=\arg\max_{v\in\mathcal{A}}\tilde s_v$, with $\sigma$ the typical fitness change per child. This balances \textit{exploiting} promising lineages and \textit{exploring} nodes with few children.
\item \textbf{Mutation.} $\mu_j^{(r)}$ diagnoses $x_{\mathrm{par}}$ and revises it into a child idea $x$ (Eq.~\ref{eq:mutation}), fitted on $\mathcal{D}_{\mathrm{train}}$.
\item \textbf{Verification and knowledge sharing.} The verifiers test $x$ (\S\ref{sec:verifiers}), return its certificate and fitness, and publish qualifying findings to $\mathcal{K}$ (\S\ref{sec:knowledge}).
\item \textbf{Child acceptance.} The unit accepts $x$ only if no previously accepted $y\in\mathcal{A}$ is at least as simple and nearly as fit: $\Cplx(y)\le\Cplx(x)$ and $F(y)\ge F(x)-\delta$, with $\Cplx$ a task-specific complexity (e.g., parameters); extra complexity must thus improve fitness by $>\delta$. It also rejects children with a confirmed constraint violation.
\item \textbf{Memory update.} \looseness=-1 The child idea $x$ joins the lineage. Accepted children become eligible parents, while rejected ones remain as negative evidence. Every scored child updates its parent's value.
\end{stages}

\noindent\textbf{Mutator agent.}\label{sec:mutator}
The mutator $\rnd{\mu_j^{(r)}=\langle m_j^{(r)},h_j^{(r)}\rangle}$, a model run in a harness, writes the child
{\setlength{\abovedisplayskip}{3pt}\setlength{\belowdisplayskip}{3pt}\setlength{\abovedisplayshortskip}{1pt}\setlength{\belowdisplayshortskip}{3pt}%
\begin{equation}
x\leftarrow\mu_j^{(r)}\bigl(x_{\mathrm{par}},\Cert(x_{\mathrm{par}}),\mathcal{L}_j^{(r-1)},\mathcal{K}_j\bigr)
\label{eq:mutation}
\end{equation}}%
\looseness=-1 from the parent, its certificate, recent lineage nodes, and the snapshot $\mathcal{K}_j$ of shared knowledge. The certificate gives \textit{local feedback} on the parent's failures; the lineage and $\mathcal{K}_j$ give \textit{global context} on what the unit has tried, why each attempt was accepted or rejected, and promising or unpromising directions. The mutator runs in a sandbox where $\mathcal{D}_{\mathrm{val}}$ and $\mathcal{D}_{\mathrm{test}}$ are inaccessible. The initial mutator is drawn from $\mathcal{G}$; the orchestrator can later change its harness or model, as we describe next.

\noindent\textbf{Orchestrator agent.}\label{sec:orchestrator}
The orchestrator $\mathcal{O}_j\in\mathcal{G}$ adapts the unit through three actions. \textsc{Retool} is considered every $P$ rounds; \textsc{Redraw} and \textsc{Restart} respond when the unit stalls, i.e., goes $P$ scored rounds without improving its best fitness.
\begin{actions}
\item \textbf{\textsc{Retool} (\textit{evolving the harness}).} Acting as a \textit{meta-mutator}, the orchestrator diagnoses traces $\mathcal{E}_j$ of recent attempts (turns, tool calls, costs, and outcomes) and proposes a harness $h'\leftarrow\mathcal{O}_j(\rnd{h_j^{(r)}},\mathcal{E}_j)$. Since a worse $h'$ would degrade every later child, a twin trial first runs the two harnesses on the same parent and mutation plan, giving children $x_i$ and $x_i'$ with $d_i:=F(x_i')-F(x_i)$ when both are scored. The unit sets $\mu_j\leftarrow\langle\rnd{m_j^{(r)}},h'\rangle$ if the $n$ scored pairs give a significantly positive mean $d_i$; otherwise, $\rnd{h_j^{(r)}}$ stays.

\item \textbf{\textsc{Redraw} (\textit{replacing the model}).} $\mathcal{K}$ tallies each model's recent fitness improvements (\S\ref{sec:knowledge}), so $\mathcal{O}_j$ can swap in a better-performing model: it draws $m'$ from this tally by Thompson sampling and, keeping the harness, sets $\mu_j\leftarrow\langle m',\rnd{h_j^{(r)}}\rangle$.
\item \textbf{\textsc{Restart} (\textit{changing the research direction}).} If the unit stalls again after a second \textsc{Redraw} with no \textsc{Verified} finding since (\S\ref{sec:knowledge}), $\mathcal{O}_j$ archives the unit's existing lineage and proposes a new seed idea from $\mathcal{K}$ to initiate a new idea tree.
\end{actions}

\subsection{Verifier Components}
\label{sec:verifiers}

The verifiers $\mathcal{V}$ \looseness=-1 evaluate each candidate idea $x$ against $x_0$ and the constraints $\mathcal{C}$. They return a certificate $\Cert(x)$ recording the outcome of every check and the evidence behind each failure, and decide which findings enter $\mathcal{K}$ (\S\ref{sec:knowledge}). Before sharing results to $\mathcal{K}$, the verifiers run four checks:
\begin{actions}
\item \textbf{Held-out scoring.} Running $x$ on $\mathcal{D}_{\mathrm{val}}$ gives $\Score(x)$ and $\Gain(x)$ (Def.~\ref{def:splits}).
\item \textbf{Constraint checking.} Each $c\in\mathcal{C}$ is tested programmatically or by LLM consensus. Confirmed violations form $\Viol(x)$, so $\Valid(x)=1-|\Viol(x)|/|\mathcal{C}|$; undecided ones form $\Undec(x)$.
\item \textbf{Out-of-distribution testing.} $\Ood(x)$ is $1$ if the score of $x$ on the held-out regime $\mathcal{D}_{\mathrm{ood}}$ is at least $\kappa$ times its score on $\mathcal{D}_{\mathrm{val}}\setminus\mathcal{D}_{\mathrm{ood}}$, else $0$.
\item \textbf{Component ablation.} Each component $t\in T(x)$ is removed in turn; its contribution on data $S\subseteq\mathcal{D}$ is the score drop $\Drop_t(x;S):=\Score(x;S)-\Score(x\setminus\{t\};S)$; its maximum over validation regimes is $\DropM_t(x):=\max_\rho\Drop_t(x;\mathcal{D}_\rho\cap\mathcal{D}_{\mathrm{val}})$ (Appendix~\ref{app:verify}).
\end{actions}

\subsection{Verified Shared Knowledge}
\label{sec:knowledge}

The shared knowledge stores the per-model gain tally used by \textsc{Redraw} (\S\ref{sec:orchestrator}) and three types of findings, $\mathcal{K}=(\mathcal{K}_{\mathrm{ver}},\mathcal{K}_{\mathrm{part}},\mathcal{K}_{\mathrm{ref}})$. Verifiers write to it, while units read it at each round's start.

\noindent\textbf{\textsc{Verified} findings.} The set $\mathcal{K}_{\mathrm{ver}}$ stores new bests $(x,\Cert(x))$. Let $x^{\mathrm{best}}$ be the highest-scoring idea in $\mathcal{K}_{\mathrm{ver}}$. A candidate $x$ is added if $\rnd{\Score(x)>\Score(x^{\mathrm{best}})+\delta}$ and it passes all checks: $\Viol(x)\cup\Undec(x)=\emptyset$, $\Ood(x)=1$, and, where ablation is supported, $\DropM_t(x)>\delta$ for each $t\in T(x)$. This raises the bar for all units.

\noindent\textbf{\textsc{Partial} findings.} The set $\mathcal{K}_{\mathrm{part}}$ holds useful components $(t,x,\Cert(x))$ for other units to reuse. Each component $t$ of a candidate $x$ with $\Valid(x)=1$ and $\Gain(x)>\delta$ enters if $\DropM_t(x)>\delta$, even when $x$ does not beat $x^{\mathrm{best}}$.

\noindent\textbf{\textsc{Refuted} findings.} The set $\mathcal{K}_{\mathrm{ref}}$ records each confirmed violation $c\in\Viol(x)$ as $(c,x,\Cert(x))$, so other units avoid them and prune search space.

\definecolor{heroMilo}{HTML}{2F7BC4}\definecolor{heroCoral}{HTML}{16A085}\definecolor{heroOE}{HTML}{E0457B}\definecolor{heroAS}{HTML}{8E5BC8}\definecolor{heroRA}{HTML}{B03A2E}
\begin{figure*}[t]
\centering
\tikzset{
  nm/.style={inner sep=0pt,font=\fontsize{7}{8}\selectfont\bfseries},
  gn/.style={inner sep=0pt,font=\fontsize{7.5}{8.5}\selectfont\bfseries,fill=white,outer sep=0pt,inner xsep=0.3mm},
  sml/.style={inner sep=0pt,font=\fontsize{6.5}{7.5}\selectfont,text=black!60},
  bsl/.style={inner sep=0pt,font=\fontsize{6.2}{7}\selectfont,text=black!62},
  key/.style={inner sep=0pt,font=\fontsize{6.5}{7.5}\selectfont,text=black!78},
  stg/.style={inner sep=0pt,font=\fontsize{5.9}{6.3}\selectfont,align=left,text=ridU!60!black},
  ptitle/.style={inner sep=0pt,anchor=base west,font=\fontsize{7.6}{9}\selectfont\bfseries},
  invl/.style={ridK,line width=0.9pt,line join=round},
  invd/.style={fill=bgK,draw=ridK,line width=0.45pt},
}%
\begin{tikzpicture}[x=1mm,y=0.94mm]
\useasboundingbox (0,0) rectangle (68.5,35.3);
\node[ptitle] at (0.3,33.2) {(a) Gain over the best baseline, by research stage};
\draw[black!45,line width=0.5pt] (38.6,3.3) -- (38.6,31.8);
\node[sml,anchor=base west,text=black!75] at (0.6,0.8) {Gain of \tikz[baseline=-0.6ex]\filldraw[invd] circle (0.75mm);\,{\color{ridK!80!black}\myTool{}} over \tikz[baseline=-0.6ex]\filldraw[fill=black!45,draw=white,line width=0.3pt] circle (0.7mm);\,the best baseline (\%)};
\fill[bgU!30] (1.3,31.8) -- (15.9,31.8) -- (15.9,26.4) -- (8.6,25.9) -- (1.3,26.4) -- cycle;
\fill[bgU!30] (1.3,25.8) -- (8.6,25.3) -- (15.9,25.8) -- (15.9,20.7) -- (8.6,20.2) -- (1.3,20.7) -- cycle;
\fill[bgU!30] (1.3,20.1) -- (8.6,19.6) -- (15.9,20.1) -- (15.9,15) -- (8.6,14.5) -- (1.3,15) -- cycle;
\fill[bgU!30] (1.3,14.4) -- (8.6,13.9) -- (15.9,14.4) -- (15.9,3.8) -- (8.6,3.3) -- (1.3,3.8) -- cycle;
\node[inner sep=0pt,anchor=east] at (4.95,29.1) {\includegraphics[height=3.2mm]{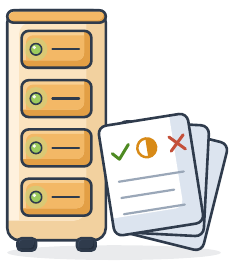}};
\node[stg,anchor=west] at (5.75,29.1) {Retrieve\\prior work};
\node[inner sep=0pt,anchor=east] at (4.95,23.1) {\includegraphics[height=3.2mm]{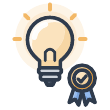}};
\node[stg,anchor=west] at (5.75,23.1) {Generate\\ideas};
\node[inner sep=0pt,anchor=east] at (4.95,17.4) {\includegraphics[height=3.2mm]{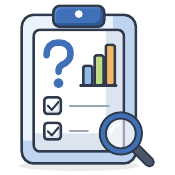}};
\node[stg,anchor=west] at (5.75,17.4) {Write\\proposals};
\node[inner sep=0pt,anchor=east] at (4.95,9.15) {\includegraphics[height=3.2mm]{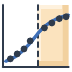}};
\node[stg,anchor=west] at (5.75,9.15) {Discover\\scientific\\laws};
\draw[black!45,line width=0.6pt,rounded corners=0.8mm,-stealth] (1.3,5.0) -- (0.5,5.0) -- (0.5,30.2) -- (1.25,30.2);
\node[nm,anchor=base west] at (16.8,29.45) {ScholarCatalyst};
\node[bsl,anchor=base west] at (16.8,27.05) {OpenCode + DS V3.2};
\draw[bgK,line width=0.55mm,line cap=round] (38.6,28.95) -- (47.81,28.95);
\filldraw[fill=black!45,draw=white,line width=0.3pt] (38.6,28.95) circle (0.55mm);
\filldraw[invd] (47.81,28.95) circle (0.75mm);
\node[gn,anchor=base west] at (49.31,29.45) {+21.9\%};
\node[sml,anchor=base west,fill=white,outer sep=0pt,inner xsep=0.3mm] at (49.31,27.05) {36.5\,$\rightarrow$\,{\color{ridK!80!black}\bfseries 44.5}};
\node[nm,anchor=base west] at (16.8,23.75) {IdeaBench};
\node[bsl,anchor=base west] at (16.8,21.35) {ResearchAgent};
\draw[bgK,line width=0.55mm,line cap=round] (38.6,23.25) -- (55.31,23.25);
\filldraw[fill=black!45,draw=white,line width=0.3pt] (38.6,23.25) circle (0.55mm);
\filldraw[invd] (55.31,23.25) circle (0.75mm);
\node[gn,anchor=base west] at (56.81,23.75) {+39.8\%};
\node[sml,anchor=base west,fill=white,outer sep=0pt,inner xsep=0.3mm] at (56.81,21.35) {18.6\,$\rightarrow$\,{\color{ridK!80!black}\bfseries 26.0}};
\node[nm,anchor=base west] at (16.8,18.05) {IdeaScientist};
\node[bsl,anchor=base west] at (16.8,15.65) {IdeaScientist-Base};
\draw[bgK,line width=0.55mm,line cap=round] (38.6,17.55) -- (41.30,17.55);
\filldraw[fill=black!45,draw=white,line width=0.3pt] (38.6,17.55) circle (0.55mm);
\filldraw[invd] (41.30,17.55) circle (0.75mm);
\node[gn,anchor=base west] at (42.80,18.05) {+6.4\%};
\node[sml,anchor=base west,fill=white,outer sep=0pt,inner xsep=0.3mm] at (42.80,15.65) {77.7\,$\rightarrow$\,{\color{ridK!80!black}\bfseries 82.7}};
\node[nm,anchor=base west] at (16.8,12.35) {EurekaBench};
\node[bsl,anchor=base west] at (16.8,9.95) {CORAL};
\draw[bgK,line width=0.55mm,line cap=round] (38.6,11.85) -- (51.60,11.85);
\filldraw[fill=black!45,draw=white,line width=0.3pt] (38.6,11.85) circle (0.55mm);
\filldraw[invd] (51.60,11.85) circle (0.75mm);
\node[gn,anchor=base west] at (53.10,12.35) {+31.0\%};
\node[sml,anchor=base west,fill=white,outer sep=0pt,inner xsep=0.3mm] at (53.10,9.95) {56.2\,$\rightarrow$\,{\color{ridK!80!black}\bfseries 73.6}};
\node[nm,anchor=base west] at (16.8,6.65) {SciLaws-Bench};
\node[bsl,anchor=base west] at (16.8,4.25) {ReAct + Grok 4.7};
\draw[bgK,line width=0.55mm,line cap=round] (38.6,6.15) -- (41.08,6.15);
\filldraw[fill=black!45,draw=white,line width=0.3pt] (38.6,6.15) circle (0.55mm);
\filldraw[invd] (41.08,6.15) circle (0.75mm);
\node[gn,anchor=base west] at (42.58,6.65) {+5.9\%};
\node[sml,anchor=base west,fill=white,outer sep=0pt,inner xsep=0.3mm] at (42.58,4.25) {59.3\,$\rightarrow$\,{\color{ridK!80!black}\bfseries 62.8}};
\ifdefined\herodebug\draw[red,line width=0.2pt] (0,0) rectangle (68.5,35.3);\fi
\end{tikzpicture}\hfill
\begin{tikzpicture}[x=1mm,y=0.94mm]
\useasboundingbox (0,0) rectangle (42.5,35.3);
\node[ptitle] at (0.3,33.2) {(b) Scientific law discovery};
\draw[black!12,line width=0.3pt] (16.46,6.5) -- (16.46,30.3);
\draw[black!12,line width=0.3pt] (24.31,6.5) -- (24.31,30.3);
\draw[black!12,line width=0.3pt] (32.17,6.5) -- (32.17,30.3);
\draw[black!12,line width=0.3pt] (40.03,6.5) -- (40.03,30.3);
\node[sml,anchor=base] at (8.60,4.0) {0};
\node[sml,anchor=base] at (16.46,4.0) {200};
\node[sml,anchor=base] at (24.31,4.0) {400};
\node[sml,anchor=base] at (32.17,4.0) {600};
\node[sml,anchor=base] at (40.03,4.0) {800};
\draw[black!12,line width=0.3pt] (8.6,12.45) -- (41.6,12.45);
\draw[black!12,line width=0.3pt] (8.6,18.40) -- (41.6,18.40);
\draw[black!12,line width=0.3pt] (8.6,24.35) -- (41.6,24.35);
\draw[black!12,line width=0.3pt] (8.6,30.30) -- (41.6,30.30);
\node[sml,anchor=east] at (7.8,6.50) {20};
\node[sml,anchor=east] at (7.8,12.45) {40};
\node[sml,anchor=east] at (7.8,18.40) {60};
\node[sml,anchor=east] at (7.8,24.35) {80};
\node[sml,anchor=east] at (7.8,30.30) {100};
\draw[black!45,line width=0.4pt] (8.6,30.3) -- (8.6,6.5) -- (41.6,6.5);
\node[sml,anchor=base,text=black!75] at (25.10,0.6) {Cost per task (units)};
\node[sml,rotate=90,anchor=center,text=black!75] at (1.9,18.40) {Cond.\ score on $\mathcal{D}_{\mathrm{val}}$ (\%)};
\draw[black!45,line width=0.7pt,line join=round,line cap=round] (10.17,9.94) -- (11.74,11.63) -- (13.31,12.53) -- (14.89,13.12) -- (16.46,13.54) -- (18.03,13.86) -- (19.60,14.12) -- (21.17,14.34) -- (22.74,14.52) -- (24.31,14.67) -- (25.89,14.81) -- (27.46,14.93) -- (29.03,15.06) -- (30.60,15.17) -- (32.17,15.28) -- (33.74,15.37) -- (35.31,15.47) -- (36.89,15.55) -- (38.46,15.63) -- (40.03,15.71);
\fill[black!45] (10.17,9.94) circle (0.36mm);
\fill[black!45] (11.74,11.63) circle (0.36mm);
\fill[black!45] (13.31,12.53) circle (0.36mm);
\fill[black!45] (14.89,13.12) circle (0.36mm);
\fill[black!45] (16.46,13.54) circle (0.36mm);
\fill[black!45] (18.03,13.86) circle (0.36mm);
\fill[black!45] (19.60,14.12) circle (0.36mm);
\fill[black!45] (21.17,14.34) circle (0.36mm);
\fill[black!45] (22.74,14.52) circle (0.36mm);
\fill[black!45] (24.31,14.67) circle (0.36mm);
\fill[black!45] (25.89,14.81) circle (0.36mm);
\fill[black!45] (27.46,14.93) circle (0.36mm);
\fill[black!45] (29.03,15.06) circle (0.36mm);
\fill[black!45] (30.60,15.17) circle (0.36mm);
\fill[black!45] (32.17,15.28) circle (0.36mm);
\fill[black!45] (33.74,15.37) circle (0.36mm);
\fill[black!45] (35.31,15.47) circle (0.36mm);
\fill[black!45] (36.89,15.55) circle (0.36mm);
\fill[black!45] (38.46,15.63) circle (0.36mm);
\fill[black!45] (40.03,15.71) circle (0.36mm);
\draw[heroOE,line width=0.7pt,line join=round,line cap=round] (10.17,13.64) -- (11.74,14.21) -- (13.31,14.53) -- (14.89,14.71) -- (16.46,14.91) -- (18.03,15.05) -- (19.60,15.17) -- (21.17,15.28) -- (22.74,15.37) -- (24.31,15.45) -- (25.89,15.52) -- (27.46,15.59) -- (29.03,15.65) -- (30.60,15.71) -- (32.17,15.77) -- (33.74,15.82) -- (35.31,15.86) -- (36.89,15.91) -- (38.46,15.95) -- (40.03,15.99);
\fill[heroOE] (10.17,14.19) -- (10.67,13.29) -- (9.67,13.29) -- cycle;
\fill[heroOE] (11.74,14.76) -- (12.24,13.86) -- (11.24,13.86) -- cycle;
\fill[heroOE] (13.31,15.08) -- (13.81,14.18) -- (12.81,14.18) -- cycle;
\fill[heroOE] (14.89,15.26) -- (15.39,14.36) -- (14.39,14.36) -- cycle;
\fill[heroOE] (16.46,15.46) -- (16.96,14.56) -- (15.96,14.56) -- cycle;
\fill[heroOE] (18.03,15.60) -- (18.53,14.70) -- (17.53,14.70) -- cycle;
\fill[heroOE] (19.60,15.72) -- (20.10,14.82) -- (19.10,14.82) -- cycle;
\fill[heroOE] (21.17,15.83) -- (21.67,14.93) -- (20.67,14.93) -- cycle;
\fill[heroOE] (22.74,15.92) -- (23.24,15.02) -- (22.24,15.02) -- cycle;
\fill[heroOE] (24.31,16.00) -- (24.81,15.10) -- (23.81,15.10) -- cycle;
\fill[heroOE] (25.89,16.07) -- (26.39,15.17) -- (25.39,15.17) -- cycle;
\fill[heroOE] (27.46,16.14) -- (27.96,15.24) -- (26.96,15.24) -- cycle;
\fill[heroOE] (29.03,16.20) -- (29.53,15.30) -- (28.53,15.30) -- cycle;
\fill[heroOE] (30.60,16.26) -- (31.10,15.36) -- (30.10,15.36) -- cycle;
\fill[heroOE] (32.17,16.32) -- (32.67,15.42) -- (31.67,15.42) -- cycle;
\fill[heroOE] (33.74,16.37) -- (34.24,15.47) -- (33.24,15.47) -- cycle;
\fill[heroOE] (35.31,16.41) -- (35.81,15.51) -- (34.81,15.51) -- cycle;
\fill[heroOE] (36.89,16.46) -- (37.39,15.56) -- (36.39,15.56) -- cycle;
\fill[heroOE] (38.46,16.50) -- (38.96,15.60) -- (37.96,15.60) -- cycle;
\fill[heroOE] (40.03,16.54) -- (40.53,15.64) -- (39.53,15.64) -- cycle;
\draw[heroMilo,line width=0.7pt,line join=round,line cap=round] (10.17,6.80) -- (11.74,7.76) -- (13.31,9.51) -- (14.89,10.29) -- (16.46,12.02) -- (18.03,13.77) -- (19.60,14.17) -- (21.17,15.58) -- (22.74,15.58) -- (24.31,16.23) -- (25.89,16.23) -- (27.46,16.23) -- (29.03,16.26) -- (30.60,16.64) -- (32.17,16.64) -- (33.74,16.64) -- (35.31,16.67) -- (36.89,16.69) -- (38.46,16.69) -- (40.03,16.69);
\fill[heroMilo] (9.77,6.40) rectangle (10.57,7.20);
\fill[heroMilo] (11.34,7.36) rectangle (12.14,8.16);
\fill[heroMilo] (12.91,9.11) rectangle (13.71,9.91);
\fill[heroMilo] (14.49,9.89) rectangle (15.29,10.69);
\fill[heroMilo] (16.06,11.62) rectangle (16.86,12.42);
\fill[heroMilo] (17.63,13.37) rectangle (18.43,14.17);
\fill[heroMilo] (19.20,13.77) rectangle (20.00,14.57);
\fill[heroMilo] (20.77,15.18) rectangle (21.57,15.98);
\fill[heroMilo] (22.34,15.18) rectangle (23.14,15.98);
\fill[heroMilo] (23.91,15.83) rectangle (24.71,16.63);
\fill[heroMilo] (25.49,15.83) rectangle (26.29,16.63);
\fill[heroMilo] (27.06,15.83) rectangle (27.86,16.63);
\fill[heroMilo] (28.63,15.86) rectangle (29.43,16.66);
\fill[heroMilo] (30.20,16.24) rectangle (31.00,17.04);
\fill[heroMilo] (31.77,16.24) rectangle (32.57,17.04);
\fill[heroMilo] (33.34,16.24) rectangle (34.14,17.04);
\fill[heroMilo] (34.91,16.27) rectangle (35.71,17.07);
\fill[heroMilo] (36.49,16.29) rectangle (37.29,17.09);
\fill[heroMilo] (38.06,16.29) rectangle (38.86,17.09);
\fill[heroMilo] (39.63,16.29) rectangle (40.43,17.09);
\draw[heroAS,line width=0.7pt,line join=round,line cap=round] (10.17,8.98) -- (11.74,12.52) -- (13.31,13.76) -- (14.89,14.01) -- (16.46,14.84) -- (18.03,15.42) -- (19.60,15.44) -- (21.17,15.44) -- (22.74,16.09) -- (24.31,16.27) -- (25.89,16.31) -- (27.46,16.34) -- (29.03,16.60) -- (30.60,16.60) -- (32.17,16.71) -- (33.74,16.80) -- (35.31,16.81) -- (36.89,16.82) -- (38.46,16.96) -- (40.03,17.08);
\fill[heroAS] (10.17,9.53) -- (10.72,8.98) -- (10.17,8.43) -- (9.62,8.98) -- cycle;
\fill[heroAS] (11.74,13.07) -- (12.29,12.52) -- (11.74,11.97) -- (11.19,12.52) -- cycle;
\fill[heroAS] (13.31,14.31) -- (13.86,13.76) -- (13.31,13.21) -- (12.76,13.76) -- cycle;
\fill[heroAS] (14.89,14.56) -- (15.44,14.01) -- (14.89,13.46) -- (14.34,14.01) -- cycle;
\fill[heroAS] (16.46,15.39) -- (17.01,14.84) -- (16.46,14.29) -- (15.91,14.84) -- cycle;
\fill[heroAS] (18.03,15.97) -- (18.58,15.42) -- (18.03,14.87) -- (17.48,15.42) -- cycle;
\fill[heroAS] (19.60,15.99) -- (20.15,15.44) -- (19.60,14.89) -- (19.05,15.44) -- cycle;
\fill[heroAS] (21.17,15.99) -- (21.72,15.44) -- (21.17,14.89) -- (20.62,15.44) -- cycle;
\fill[heroAS] (22.74,16.64) -- (23.29,16.09) -- (22.74,15.54) -- (22.19,16.09) -- cycle;
\fill[heroAS] (24.31,16.82) -- (24.86,16.27) -- (24.31,15.72) -- (23.76,16.27) -- cycle;
\fill[heroAS] (25.89,16.86) -- (26.44,16.31) -- (25.89,15.76) -- (25.34,16.31) -- cycle;
\fill[heroAS] (27.46,16.89) -- (28.01,16.34) -- (27.46,15.79) -- (26.91,16.34) -- cycle;
\fill[heroAS] (29.03,17.15) -- (29.58,16.60) -- (29.03,16.05) -- (28.48,16.60) -- cycle;
\fill[heroAS] (30.60,17.15) -- (31.15,16.60) -- (30.60,16.05) -- (30.05,16.60) -- cycle;
\fill[heroAS] (32.17,17.26) -- (32.72,16.71) -- (32.17,16.16) -- (31.62,16.71) -- cycle;
\fill[heroAS] (33.74,17.35) -- (34.29,16.80) -- (33.74,16.25) -- (33.19,16.80) -- cycle;
\fill[heroAS] (35.31,17.36) -- (35.86,16.81) -- (35.31,16.26) -- (34.76,16.81) -- cycle;
\fill[heroAS] (36.89,17.37) -- (37.44,16.82) -- (36.89,16.27) -- (36.34,16.82) -- cycle;
\fill[heroAS] (38.46,17.51) -- (39.01,16.96) -- (38.46,16.41) -- (37.91,16.96) -- cycle;
\fill[heroAS] (40.03,17.63) -- (40.58,17.08) -- (40.03,16.53) -- (39.48,17.08) -- cycle;
\draw[heroCoral,line width=0.7pt,line join=round,line cap=round] (10.17,7.44) -- (11.74,8.30) -- (13.31,9.71) -- (14.89,11.03) -- (16.46,11.84) -- (18.03,11.84) -- (19.60,12.25) -- (21.17,13.11) -- (22.74,14.67) -- (24.31,15.44) -- (25.89,15.79) -- (27.46,15.80) -- (29.03,15.80) -- (30.60,15.93) -- (32.17,16.39) -- (33.74,16.39) -- (35.31,16.47) -- (36.89,17.02) -- (38.46,17.02) -- (40.03,17.15);
\fill[heroCoral] (10.17,7.44) circle (0.36mm);
\fill[heroCoral] (11.74,8.30) circle (0.36mm);
\fill[heroCoral] (13.31,9.71) circle (0.36mm);
\fill[heroCoral] (14.89,11.03) circle (0.36mm);
\fill[heroCoral] (16.46,11.84) circle (0.36mm);
\fill[heroCoral] (18.03,11.84) circle (0.36mm);
\fill[heroCoral] (19.60,12.25) circle (0.36mm);
\fill[heroCoral] (21.17,13.11) circle (0.36mm);
\fill[heroCoral] (22.74,14.67) circle (0.36mm);
\fill[heroCoral] (24.31,15.44) circle (0.36mm);
\fill[heroCoral] (25.89,15.79) circle (0.36mm);
\fill[heroCoral] (27.46,15.80) circle (0.36mm);
\fill[heroCoral] (29.03,15.80) circle (0.36mm);
\fill[heroCoral] (30.60,15.93) circle (0.36mm);
\fill[heroCoral] (32.17,16.39) circle (0.36mm);
\fill[heroCoral] (33.74,16.39) circle (0.36mm);
\fill[heroCoral] (35.31,16.47) circle (0.36mm);
\fill[heroCoral] (36.89,17.02) circle (0.36mm);
\fill[heroCoral] (38.46,17.02) circle (0.36mm);
\fill[heroCoral] (40.03,17.15) circle (0.36mm);
\draw[invl] (24.04,19.81) -- (39.48,22.11);
\filldraw[invd] (24.04,19.81) circle (0.75mm);
\filldraw[invd] (39.48,22.11) circle (0.75mm);
\draw[ridK!80!black,line width=0.5pt,-stealth,shorten <=0.75mm,shorten >=1.05mm] (24.04,16.27) -- (24.04,19.81);
\node[key,anchor=west,text=ridK!80!black,font=\fontsize{6.4}{7.4}\selectfont\bfseries] at (24.94,18.04) {+11.9};
\draw[ridK!80!black,line width=0.5pt,-stealth,shorten <=0.75mm,shorten >=1.05mm] (39.48,17.07) -- (39.48,22.11);
\node[key,anchor=east,text=ridK!80!black,font=\fontsize{6.4}{7.4}\selectfont\bfseries] at (38.58,19.59) {+17.0};
\node[key,anchor=south,text=ridK!80!black,font=\fontsize{6.8}{8}\selectfont\bfseries,rotate=8.47] at (31.76,21.71) {\myTool{}};
\draw[heroCoral,line width=0.7pt] (9.70,29.15) -- (11.70,29.15);
\fill[heroCoral] (10.70,29.15) circle (0.42mm);
\node[key,anchor=west,font=\fontsize{5.8}{6.6}\selectfont] at (12.10,29.15) {CORAL};
\draw[heroAS,line width=0.7pt] (9.70,27.15) -- (11.70,27.15);
\fill[heroAS] (10.70,27.70) -- (11.25,27.15) -- (10.70,26.60) -- (10.15,27.15) -- cycle;
\node[key,anchor=west,font=\fontsize{5.8}{6.6}\selectfont] at (12.10,27.15) {Agentic Scientist};
\draw[heroMilo,line width=0.7pt] (9.70,25.15) -- (11.70,25.15);
\fill[heroMilo] (10.30,24.75) rectangle (11.10,25.55);
\node[key,anchor=west,font=\fontsize{5.8}{6.6}\selectfont] at (12.10,25.15) {MILO};
\draw[heroOE,line width=0.7pt] (27.40,29.15) -- (29.40,29.15);
\fill[heroOE] (28.40,29.70) -- (28.90,28.80) -- (27.90,28.80) -- cycle;
\node[key,anchor=west,font=\fontsize{5.8}{6.6}\selectfont] at (29.80,29.15) {OpenEvolve};
\draw[black!45,line width=0.7pt] (27.40,27.15) -- (29.40,27.15);
\fill[black!45] (28.40,27.15) circle (0.42mm);
\node[key,anchor=west,font=\fontsize{5.8}{6.6}\selectfont] at (29.80,27.15) {Agents \rid{1}\unkern--\rid{14}\unkern};
\ifdefined\herodebug\draw[red,line width=0.2pt] (0,0) rectangle (42.5,35.3);\fi
\end{tikzpicture}\hfill
\begin{tikzpicture}[x=1mm,y=0.94mm]
\useasboundingbox (0,0) rectangle (42.5,35.3);
\node[ptitle] at (0.3,33.2) {(c) Idea generation};
\draw[black!12,line width=0.3pt] (15.20,6.5) -- (15.20,30.3);
\draw[black!12,line width=0.3pt] (21.80,6.5) -- (21.80,30.3);
\draw[black!12,line width=0.3pt] (28.40,6.5) -- (28.40,30.3);
\draw[black!12,line width=0.3pt] (35.00,6.5) -- (35.00,30.3);
\draw[black!12,line width=0.3pt] (41.60,6.5) -- (41.60,30.3);
\node[sml,anchor=base] at (8.60,4.0) {0};
\node[sml,anchor=base] at (15.20,4.0) {100};
\node[sml,anchor=base] at (21.80,4.0) {200};
\node[sml,anchor=base] at (28.40,4.0) {300};
\node[sml,anchor=base] at (35.00,4.0) {400};
\node[sml,anchor=base] at (41.60,4.0) {500};
\draw[black!12,line width=0.3pt] (8.6,13.30) -- (41.6,13.30);
\draw[black!12,line width=0.3pt] (8.6,20.10) -- (41.6,20.10);
\draw[black!12,line width=0.3pt] (8.6,26.90) -- (41.6,26.90);
\node[sml,anchor=east] at (7.8,6.50) {0};
\node[sml,anchor=east] at (7.8,13.30) {40};
\node[sml,anchor=east] at (7.8,20.10) {80};
\node[sml,anchor=east] at (7.8,26.90) {120};
\draw[black!45,line width=0.4pt] (8.6,30.3) -- (8.6,6.5) -- (41.6,6.5);
\node[sml,anchor=base,text=black!75] at (25.10,0.6) {Model calls per paper};
\node[sml,rotate=90,anchor=center,text=black!75,font=\fontsize{6}{7}\selectfont] at (1.9,17.40) {Distinct novel ideas / paper};
\draw[black!55,line width=0.75pt,dash pattern=on 0.5pt off 0.9pt,line cap=round] (11.90,7.30) -- (24.57,8.00);
\fill[black!55] (11.90,7.30) circle (0.45mm);
\fill[black!55] (24.57,8.00) circle (0.45mm);
\draw[heroMilo,line width=0.75pt,dash pattern=on 2.0pt off 1.1pt] (12.30,8.78) -- (25.10,14.69) -- (40.87,20.68);
\filldraw[fill=heroMilo,draw=white,line width=0.3pt] (11.80,8.28) rectangle (12.80,9.28);
\filldraw[fill=heroMilo,draw=white,line width=0.3pt] (24.60,14.19) rectangle (25.60,15.19);
\filldraw[fill=heroMilo,draw=white,line width=0.3pt] (40.37,20.18) rectangle (41.37,21.18);
\draw[ridK!80!black,line width=0.5pt,-stealth,shorten <=0.9mm,shorten >=0.3mm] (25.10,14.69) -- (14.87,14.69);
\draw[invl] (11.57,10.92) -- (14.87,16.97) -- (27.94,28.63);
\filldraw[invd] (11.57,10.92) circle (0.75mm);
\filldraw[invd] (14.87,16.97) circle (0.75mm);
\filldraw[invd] (27.94,28.63) circle (0.75mm);
\node[key,anchor=west,text=ridK!80!black,font=\fontsize{6.8}{8}\selectfont\bfseries] at (29.34,28.63) {\myTool{}};
\draw[heroMilo,line width=0.75pt,dash pattern=on 2.0pt off 1.1pt] (27.90,25.40) -- (29.90,25.40);
\filldraw[fill=heroMilo,draw=white,line width=0.3pt] (28.45,24.95) rectangle (29.35,25.85);
\node[key,anchor=west,font=\fontsize{5.7}{6.5}\selectfont] at (30.30,25.40) {MILO};
\draw[black!55,line width=0.75pt,dash pattern=on 0.5pt off 0.9pt,line cap=round] (27.90,23.45) -- (29.90,23.45);
\fill[black!55] (28.90,23.45) circle (0.42mm);
\node[key,anchor=west,font=\fontsize{5.7}{6.5}\selectfont] at (30.30,23.45) {Best-of-$N$};
\node[key,anchor=north west,align=left,font=\fontsize{6.2}{7}\selectfont] at (9.6,30.10) {20--62\% fewer\\calls than MILO};
\ifdefined\herodebug\draw[red,line width=0.2pt] (0,0) rectangle (42.5,35.3);\fi
\end{tikzpicture}
\vspace{-2.5mm}
\caption{\textbf{\myTool{} leads at every research stage and keeps improving with search.} \looseness=-1 (a)~Gain over the best baseline on the respective primary metric (\%). (b)~EurekaBench validation conditioned score vs.\ cost per task, in units of 10\,USD; baselines from Table~\ref{tab:eb}. (c)~IdeaBench distinct novel ideas vs.\ model calls per paper.}
\label{fig:cycle}
\end{figure*}

\section{Experiments}
\label{sec:experiments}

\noindent\textbf{Benchmarks and metrics.}\label{sec:experiments:metrics} \looseness=-1 We evaluate \myTool{} at four stages of the research cycle: (i) \emph{retrieving prior work that inspires new ideas} on ScholarCatalyst~\citep{kim2026scholarcatalyst} (894 queries), (ii) \emph{generating ideas} on IdeaBench~\citep{guo2024ideabench} (736 test papers), (iii) \emph{writing grounded proposals} on IdeaScientist~\citep{liu2026ideascientist} (277 research problems), and (iv) \emph{discovering scientific laws} on EurekaBench~\citep{geng2026eurekabench} (16 CPU-only tasks) and SciLaws-Bench~\citep{huang2026scilaws} (38 tasks). Primary metrics follow each stage's goal: (i) retrieval quality, the mean of nDCG@20 and Recall@$k$; (ii) novelty, distinct novel ideas per paper rated $\geq$7/10 by an expert-calibrated judge; (iii) novelty and quality, their combined score; and (iv) constraint-aware fit, EurekaBench's conditioned score (CS) of predictive accuracy (PA) and scientific insight (SI), zero unless the law passes all of the task's scientific-constraint tests, and SciLaws-Bench's held-out fit $S_N$ (see Appendices~\ref{app:setup:benchmarks}--\ref{app:metrics}).

\noindent\textbf{Baselines and protocol.}\label{sec:experiments:setup} We compare with frontier AI agents (OpenHands~\citep{openhands2024}, OpenCode~\citep{opencode2025}, Goose~\citep{goose2025}, SciLaws-Bench's ReAct) with up to five models, Best-of-$N$, SoTA evolutionary search and AI co-scientists (e.g., GEPA~\citep{agrawal2026gepa}, MILO, CORAL), and SoTA task-specific systems. Evolution sees only $\mathcal{D}_{\mathrm{train}}$ and $\mathcal{D}_{\mathrm{val}}$; the best candidate, chosen on $\mathcal{D}_{\mathrm{val}}$, is then tested on $\mathcal{D}_{\mathrm{test}}$ (details in Appendix~\ref{app:setup:repro}).

\findingw{RQ1: Can one framework serve multiple stages of the research cycle?\label{sec:experiments:results}}
\begin{wraptable}{r}{\wSC}
\vspace{\wSCtop}
\centering
\caption{\textbf{ScholarCatalyst} (894 queries; primary: \textit{Mean}). \best{Bold}/\nd{underline}: best/second.}
\label{tab:sc}
\footnotesize\setlength{\aboverulesep}{0.6pt}\setlength{\belowrulesep}{1pt}\renewcommand{\arraystretch}{\rsSC}
\fitat{\tsc}{10}{%
\begin{tabular}{@{}l !{\color{black!22}\vrule} N N N N !{\color{black!22}\vrule} N @{}}
\toprule
 & \multicolumn{4}{c!{\color{black!22}\vrule}}{\mh{All 894 queries (\%)$\uparrow$}} & \mh{Mean} \\
\cmidrule(lr){2-5}
\textbf{Method} & \mh{nDCG@20} & \mh{R@5} & \mh{R@20} & \mh{R@100} & \mh{of 8$\uparrow$} \\
\midrule
\grp{6}{Retrievers without a language model}\\
\ind\rid{1}BM25 \citeyearpar{robertson2009bm25} & 19.6 & 11.9 & 25.2 & 44.0 & 24.2 \\
\ind\rid{2}Cohere Embed v4 \citeyearpar{cohere2025embed4} & 30.1 & 19.2 & 37.7 & 60.8 & 35.4 \\
\ind\rid{3}Cohere v4 + BM25 fusion & 28.3 & 18.2 & 35.7 & 60.9 & 34.1 \\
\ind\rid{4}Cohere v4 + Rerank 3.5 \citeyearpar{cohere2024rerank35} & 29.5 & 18.9 & \nd{39.0} & 60.8 & 35.6 \\
\grp{6}{DeepSeek V3.2 agents (Cohere v4 retriever)}\\
\ind\rid{5}Tool-calling Agent \citeyearpar{kim2026scholarcatalyst} & 28.7 & 18.8 & 36.0 & 60.7 & 34.5 \\
\ind\rid{6}+ real metadata \citeyearpar{kim2026scholarcatalyst} & 28.4 & 18.9 & 36.1 & 60.8 & 34.7 \\
\ind\rid{7}Deep-research Agent \citeyearpar{kim2026scholarcatalyst} & 30.2 & 20.0 & 36.3 & 61.0 & 35.2 \\
\ind\rid{8}OpenCode \citeyearpar{opencode2025} & \nd{31.6} & \nd{21.7} & 38.3 & \nd{61.7} & \nd{36.5} \\
\ind\rid{9}OpenHands \citeyearpar{openhands2024} & 30.7 & 21.4 & 38.3 & 59.8 & 36.3 \\
\ind\rid{10}Seed agent (initialization) & 27.3 & 19.3 & 33.1 & 58.4 & 33.1 \\
\rowcolor{tbOurs}\oursL{\bandstrut\ind\rid{11}\textbf{\myTool{}}} & \best{39.1} & \best{26.9} & \best{49.5} & \best{69.9} & \oursR{\best{44.5}} \\
\bottomrule
\end{tabular}}
\vspace{\wSCbot}
\end{wraptable}
\noindent \myTool{} ranks first on the primary metric of all five benchmarks, 6--40\% above the strongest baseline (Fig.~\ref{fig:cycle}a). \textit{SoTA frameworks, by contrast, are inconsistent across stages}: among evolutionary baselines, CORAL leads on \mbox{EurekaBench} (56.2\% CS) but trails on IdeaBench (2.7 ideas), and OpenEvolve does the opposite (14.7 ideas; 34.0\% CS), whereas \myTool{} leads on both. In retrieval (\textbf{ScholarCatalyst}), \myTool{} evolves its seed retriever from 33.1\% to 44.5\%, versus 36.5\% for the best agent, OpenCode (Table~\ref{tab:sc}). In idea generation (\textbf{IdeaBench}), it returns 26.0 distinct, judge-rated novel ideas per paper, versus 18.6 for ResearchAgent, with 82\% fewer calls (Table~\ref{tab:ib}). In proposal writing (\textbf{IdeaScientist}), \myTool{} scores 82.7\%, vs.\ 77.7\% for IdeaScientist-Base and 74.6\% for IdeaScientist-Trained, with the highest novelty (82.3\% vs.\ at most 69.8\%; Table~\ref{tab:is}). In scientific law discovery (\mbox{\textbf{EurekaBench}}, \mbox{\textbf{SciLaws-Bench}}), it leads by 31.0\% and 5.9\%.

\begin{table*}[t]
\centering
\begin{minipage}[t]{\wEB}\vspace{0pt}
\centering
\caption{\textbf{EurekaBench} (16 tasks; primary: CS). Best-of-$N$: best on $\mathcal{D}_{\mathrm{val}}$, tested on $\mathcal{D}_{\mathrm{test}}$. \best{Bold}/\nd{underline}: best/second.}
\label{tab:eb}
\footnotesize\setlength{\aboverulesep}{0.6pt}\setlength{\belowrulesep}{1pt}\renewcommand{\arraystretch}{\rsEB}
\fitat{\tsc}{8}{%
\begin{tabular}{@{}l !{\color{black!22}\vrule} N N N !{\color{black!22}\vrule} N @{}}
\toprule
\textbf{Method} & \mhh{Constraint}{fails (SC)$\downarrow$} & \mhh{Predictive}{acc.\ (PA, \%)$\uparrow$} & \mhh{Scientific}{insights (SI, \%)$\uparrow$} & \mhh{Conditioned}{score (CS, \%)$\uparrow$} \\
\midrule
\grp{5}{Frontier AI agents}\\
\subgrp{5}{OpenHands {\normalfont(\citeyear{openhands2024}; Best-of-20)}}\\
\ind\rid{1}DeepSeek V3.2 & 12/16 & 21.6 & 6.7 & 3.6 \\
\ind\rid{2}MiniMax M2.5 & 13/16 & 23.9 & 8.7 & 5.9 \\
\ind\rid{3}Kimi K3 & 6/16 & 49.9 & 35.1 & 30.9 \\
\ind\rid{4}Grok 4.7 & 8/16 & 53.1 & 38.6 & 28.0 \\
\subgrp{5}{OpenCode {\normalfont(\citeyear{opencode2025}; Best-of-20)}}\\
\ind\rid{5}DeepSeek V3.2 & 11/16 & 19.4 & 7.1 & 6.6 \\
\ind\rid{6}MiniMax M2.5 & 12/16 & 29.4 & 7.8 & 6.1 \\
\ind\rid{7}Kimi K3 & 12/16 & 34.1 & 30.6 & 15.2 \\
\ind\rid{8}GLM 5.3 & 11/16 & 42.6 & 29.1 & 20.5 \\
\ind\rid{9}Grok 4.7 & 7/16 & 39.0 & 42.1 & 31.4 \\
\subgrp{5}{Goose {\normalfont(\citeyear{goose2025}; Best-of-20)}}\\
\ind\rid{10}DeepSeek V3.2 & 10/16 & 19.6 & 9.9 & 8.8 \\
\ind\rid{11}MiniMax M2.5 & 14/16 & 28.8 & 9.3 & 9.0 \\
\ind\rid{12}Kimi K3 & 11/16 & 35.2 & 26.8 & 14.2 \\
\ind\rid{13}GLM 5.3 & 9/16 & 42.3 & 28.6 & 23.3 \\
\ind\rid{14}Grok 4.7 & 4/16 & 44.4 & 44.1 & 37.2 \\
\arrayrulecolor{black!30}\specialrule{0.3pt}{0pt}{-0.3pt}\arrayrulecolor{black}
\ind\rid{15}Best of \rid{1}\unkern--\rid{14}\unkern & 5/16 & 46.9 & 44.5 & 38.1 \\
\grp{5}{Evolutionary search \& co-scientists}\\
\ind\rid{16}OpenEvolve \citeyearpar{sharma2025openevolve} & 6/16 & 59.1 & 31.3 & 34.0 \\
\ind\rid{17}GEPA \citeyearpar{agrawal2026gepa} & 4/16 & 53.9 & 38.4 & 38.5 \\
\ind\rid{18}EvoX \citeyearpar{liu2026evox} & 3/16 & 56.0 & 29.1 & 36.2 \\
\ind\rid{19}MILO \citeyearpar{jana2026milo} & 4/16 & 78.8 & 44.2 & 47.7 \\
\ind\rid{20}CORAL \citeyearpar{qu2026coral} & 3/16 & \nd{81.7} & \nd{50.3} & \nd{56.2} \\
\ind\rid{21}Agentic Scientist \citeyearpar{lange2026agentic} & \nd{1/16} & 59.8 & 42.1 & 47.8 \\
\ind\rid{22}InternAgent-1.5 \citeyearpar{feng2026internagent} & 5/16 & 65.2 & 34.0 & 36.3 \\
\midrule
\rowcolor{tbOurs}\oursL{\bandstrut\ind\rid{23}\textbf{\myTool{}}} & \best{0/16} & \best{85.1} & \best{67.0} & \oursR{\best{73.6}} \\
\bottomrule
\end{tabular}}
\end{minipage}\hfill
\begin{minipage}[t]{\wSLB}\vspace{0pt}
\centering
\caption{\textbf{SciLaws-Bench} (38 tasks; primary: $S_N$). $>$Ref: beats the published law. \best{Bold}/\nd{underline}: best/second.}
\label{tab:slb}
\footnotesize\setlength{\aboverulesep}{0.6pt}\setlength{\belowrulesep}{1pt}\renewcommand{\arraystretch}{\rsSLB}\renewcommand{\numsz}{\fontsize{10.4}{10.4}\selectfont}
\fitat{\tsc}{12}{%
\begin{tabular}{@{}l !{\color{black!22}\vrule} N N N c !{\color{black!22}\vrule} N !{\color{black!22}\vrule} N @{}}
\toprule
 & \multicolumn{4}{c!{\color{black!22}\vrule}}{\mh{Held-out predictive fit ($S_N$, \%)$\uparrow$}} & \multirow{2}{*}[-0.2ex]{\mhh{Beats law}{(${>}$Ref)$\uparrow$}} & \multirow{2}{*}[-0.2ex]{\mhh{Validity}{($S_V$, \%)$\uparrow$}} \\
\cmidrule(lr){2-5}
\textbf{Method} & \mh{Single} & \mh{Multi} & \mh{All} & \mh{95\% CI} & & \\
\midrule
\grp{7}{Frontier AI agents}\\
\subgrp{7}{SciLaws-Bench's~{\normalfont\citep{huang2026scilaws}} ReAct harness {\normalfont(Best-of-2)}}\\
\ind\rid{1}DeepSeek V3.2 & 47.6 & 42.8 & 44.8 & {\scriptsize[38.8, 50.2]} & 14/38 & 73.4 \\
\ind\rid{2}MiniMax M2.5 & 45.4 & 44.1 & 44.6 & {\scriptsize[38.1, 50.6]} & 16/38 & 71.9 \\
\ind\rid{3}Kimi K3 & 59.3 & 51.3 & 54.7 & {\scriptsize[49.7, 59.3]} & 22/38 & \nd{77.9} \\
\ind\rid{4}GLM 5.3 & 56.7 & 51.7 & 53.8 & {\scriptsize[49.1, 58.1]} & 23/38 & 75.1 \\
\ind\rid{5}Grok 4.7 & 59.4 & 55.1 & 56.9 & {\scriptsize[53.9, 60.2]} & 22/38 & 76.7 \\
\subgrp{7}{SciLaws-Bench's ReAct harness {\normalfont(Best-of-50)}}\\
\ind\rid{6}DeepSeek V3.2 & 53.5 & 56.0 & 55.0 & {\scriptsize[50.7, 59.0]} & 23/38 & 74.4 \\
\ind\rid{7}MiniMax M2.5 & 50.8 & 58.8 & 55.4 & {\scriptsize[50.1, 60.0]} & 25/38 & 71.5 \\
\ind\rid{8}Kimi K3 & \nd{59.8} & 57.8 & 58.6 & {\scriptsize[55.0, 62.5]} & 26/38 & 74.0 \\
\ind\rid{9}GLM 5.3 & 59.6 & 58.8 & 59.1 & {\scriptsize[55.3, 63.2]} & 24/38 & 70.0 \\
\ind\rid{10}Grok 4.7 & \best{60.4} & 58.4 & \nd{59.3} & {\scriptsize[56.2, 62.6]} & \nd{28/38} & \best{79.3} \\
\grp{7}{Evolutionary search \& co-scientists}\\
\ind\rid{11}OpenEvolve \citeyearpar{sharma2025openevolve} & 44.5 & 57.4 & 52.0 & {\scriptsize[46.8, 56.7]} & 18/38 & 70.9 \\
\ind\rid{12}GEPA \citeyearpar{agrawal2026gepa} & 54.1 & 54.7 & 54.4 & {\scriptsize[49.9, 58.8]} & 18/38 & 73.4 \\
\ind\rid{13}EvoX \citeyearpar{liu2026evox} & 36.5 & 54.9 & 47.2 & {\scriptsize[40.8, 52.8]} & 14/38 & 75.9 \\
\ind\rid{14}MILO \citeyearpar{jana2026milo} & 55.2 & \nd{61.4} & 58.9 & {\scriptsize[54.1, 63.4]} & 25/38 & 74.2 \\
\ind\rid{15}CORAL \citeyearpar{qu2026coral} & 54.6 & 60.1 & 57.9 & {\scriptsize[52.8, 62.7]} & 24/38 & 75.3 \\
\ind\rid{16}Agentic Scientist \citeyearpar{lange2026agentic} & 50.7 & 55.8 & 53.8 & {\scriptsize[48.9, 58.6]} & 21/38 & 76.1 \\
\ind\rid{17}InternAgent-1.5 \citeyearpar{feng2026internagent} & 48.9 & 56.2 & 53.3 & {\scriptsize[48.4, 58.2]} & 20/38 & 72.8 \\
\midrule
\rowcolor{tbOurs}\oursL{\bandstrut\ind\rid{18}\textbf{\myTool{}}} & 58.6 & \best{65.8} & \best{62.8} & {\scriptsize[57.1, 67.9]} & \best{31/38} & \oursR{77.1} \\
\bottomrule
\end{tabular}}
\par\vspace{\gapIS}
\caption{\textbf{IdeaScientist} (277 problems; primary: \textit{Combined}). Rows 1--2 from \citet{liu2026ideascientist}.}
\label{tab:is}
\footnotesize\setlength{\aboverulesep}{0.6pt}\setlength{\belowrulesep}{1pt}
\fitat{\tsc}{12}{%
\begin{tabular}{@{}l !{\color{black!22}\vrule} N N N N !{\color{black!22}\vrule} N !{\color{black!22}\vrule} N @{}}
\toprule
 & \multicolumn{4}{c!{\color{black!22}\vrule}}{\mh{Novelty (\%)}} & \multirow{2}{*}[-0.2ex]{\mhh{Proposal}{quality (\%)}} & \multirow{2}{*}[-0.2ex]{\mhh{Combined}{score (\%)$\uparrow$}} \\[-2pt]
\cmidrule(lr){2-5}\noalign{\vskip-2.5pt}
\textbf{Method} & \mh{all} & \mh{cutoff} & \mh{in-domain} & \mh{mean} & & \\
\midrule
\ind IdeaScientist-Base & \best{77.6} & \best{77.0} & 54.9 & \nd{69.8} & \best{85.7} & \nd{77.7} \\
\ind IdeaScientist-Trained & 67.3 & 68.3 & \nd{70.9} & 68.8 & 80.4 & 74.6 \\
\rowcolor{tbOurs}\oursL{\bandstrut\ind\textbf{\myTool{}}} & \nd{76.4} & \nd{76.7} & \best{93.9} & \best{82.3} & \nd{83.0} & \oursR{\best{82.7}} \\
\bottomrule
\end{tabular}}
\end{minipage}
\end{table*}

\begin{table*}[t]
\centering
\captionsetup{justification=raggedright}
\caption{\textbf{IdeaBench} (736 test papers; App.~\ref{app:metrics}). \textit{Distinct novel}: ideas per paper rated $\geq$8 or $\geq$7 (primary). \best{Bold}/\nd{underline}: best/second per setting.}
\label{tab:ib}
\footnotesize\setlength{\aboverulesep}{0.6pt}\setlength{\belowrulesep}{1pt}
\fitat{\tsc}{20}{%
\begin{tabular}{@{}l !{\color{black!22}\vrule} R !{\color{black!22}\vrule}
  N N !{\color{black!22}\vrule}
  N N N !{\color{black!22}\vrule} N N !{\color{black!22}\vrule}
  N N @{}}
\toprule
 & & \multicolumn{2}{c!{\color{black!22}\vrule}}{\mh{Novelty \& breadth}}
 & \multicolumn{3}{c!{\color{black!22}\vrule}}{\mh{Insight Score$\uparrow$}}
 & \multicolumn{2}{c!{\color{black!22}\vrule}}{\mh{Similarity to paper}}
 & \multicolumn{2}{c}{\mh{Distinct novel$\uparrow$}} \\
\cmidrule(lr){3-4}\cmidrule(lr){5-7}\cmidrule(lr){8-9}\cmidrule(lr){10-11}
\textbf{Method} & \mhh{Calls}{/paper}
 & \mhh{Max}{Novelty$\uparrow$} & \mhh{Non-duplicate}{Rate$\uparrow$}
 & \mhh{Novelty}{published}
 & \mhh{Novelty}{disjoint}
 & \mh{Feasibility} & \mh{Semantic}
 & \mh{Overlap}
 & \mh{${\geq}8$} & \mh{${\geq}7$} \\
\midrule
\grp{11}{27-idea setting}\\
\ind\rid{1}Single prompt
 & 3 & 7.75 & 42.4 & 94.8 & 91.6 & 1.5 & 54.3 & 3.87 & 1.0 & 1.2 \\
\ind\rid{2}Over-generate + rerank
 & 21 & 8.09 & 21.3 & 97.0 & 95.2 & 1.6 & 54.6 & 3.53 & 3.9 & 4.1 \\
\ind\rid{3}Best-of-$N$, $N{=}26$
 & 50 & 7.96 & 21.8 & 96.5 & 94.1 & 8.6 & 55.0 & 3.77 & 2.2 & 4.7 \\
\ind\rid{4}GEPA \citeyearpar{agrawal2026gepa}
 & 56 & 8.00 & 40.8 & 97.1 & 96.5 & \nd{32.2} & 53.7 & 2.99 & 8.3 & 12.3 \\
\ind\rid{5}OpenEvolve \citeyearpar{sharma2025openevolve}
 & 56 & \nd{8.63} & 70.6 & \best{98.7} & \best{98.6} & 5.3 & 52.8 & 2.68 & \nd{13.5} & 14.7 \\
\ind\rid{6}EvoX \citeyearpar{liu2026evox}
 & 56 & 8.13 & 22.6 & 97.0 & 94.6 & 5.8 & 54.1 & 3.47 & 7.7 & 8.2 \\
\ind\rid{7}CORAL \citeyearpar{qu2026coral}
 & 41 & 8.08 & 30.5 & \nd{98.4} & 96.4 & \best{55.3} & 50.5 & 2.55 & 2.5 & 2.7 \\
\ind\rid{8}ResearchAgent \citeyearpar{baek2025researchagent}
 & 247 & 7.96 & \nd{97.9} & 95.6 & 95.2 & 25.5 & 54.4 & 3.15 & 5.3 & \nd{18.6} \\
\ind\rid{9}MILO \citeyearpar{jana2026milo}, 3 islands
 & 56 & 8.44 & 52.0 & 98.3 & \nd{98.1} & 13.8 & 54.3 & 3.25 & 10.1 & 13.4 \\
\rowcolor{tbOurs}\oursL{\bandstrut\ind\rid{10}\textbf{\myTool{}}}
 & 45 & \best{8.88} & \best{98.3} & 96.8 & 97.4 & 20.8 & 51.7 & 2.25 & \best{18.4} & \oursR{\best{26.0}} \\
\grp{11}{122-idea setting}\\
\ind\rid{11}Best-of-$N$, $N{=}122$
 & 242 & \nd{8.02} & 8.2 & 94.9 & 92.7 & 9.7 & 54.9 & 3.72 & 4.0 & 8.8 \\
\ind\rid{12}MILO, 15 islands
 & 250 & \best{8.82} & \nd{40.9} & \best{98.5} & \best{98.4} & \nd{17.1} & 54.4 & 3.36 & \best{32.5} & \nd{48.2} \\
\rowcolor{tbOurs}\oursL{\bandstrut\ind\rid{13}\textbf{\myTool{}}}
 & 95 & 7.99 & \best{59.2} & \nd{96.0} & \nd{97.1} & \best{32.9} & 53.3 & 2.60 & \nd{16.7} & \oursR{\best{61.6}} \\
\grp{11}{241-idea setting}\\
\ind\rid{14}MILO, 30 islands
 & 489 & \best{8.93} & \nd{35.8} & \best{97.9} & \best{97.9} & \nd{17.4} & 54.4 & 3.35 & \best{59.2} & \nd{83.4} \\
\rowcolor{tbOurs}\oursL{\bandstrut\ind\rid{15}\textbf{\myTool{}}}
 & 293 & \nd{8.08} & \best{67.9} & \nd{94.8} & \nd{96.1} & \best{25.8} & 53.2 & 2.42 & \nd{52.0} & \oursR{\best{130.2}} \\
\bottomrule
\end{tabular}}
\end{table*}

\finding{RQ2: Are the discovered scientific laws both novel and consistent with domain constraints?}\pullup On \textbf{EurekaBench}, \myTool{} alone passes every scientific-constraint test, and its laws carry far more insight than those of CORAL, the strongest baseline (67.0\% vs.\ 50.3\% SI; 73.6\% vs.\ 56.2\% CS; Table~\ref{tab:eb}). On \mbox{\textbf{SciLaws-Bench}}, \myTool{} beats the published law on 31 of 38 tasks and reaches 62.8\% $S_N$, against 28 and 59.3\% for the best baseline (Table~\ref{tab:slb}); its edge lies in tasks that hold out entire groups (65.8\% vs.\ at most 61.4\%). Its validity $S_V$ (77.1\%) also exceeds that of 15 of the 17 baselines. \myTool{} thus discovers scientific laws that are both novel and consistent with domain constraints.\stoppullup

\finding{RQ3: How does performance scale with the search budget?}\myTool{} turns extra budget into steady gains, whereas SoTA agents and evolutionary methods level off. On \textbf{EurekaBench} validation at equal cost (Fig.~\ref{fig:cycle}b), \myTool{} reaches 64.7\% CS after 10 rounds, against at most 52.8\% for the SoTA evolutionary methods and 47.4\% for the best of agents \rid{1}\unkern--\rid{14}\unkern; after 20 rounds, it reaches 72.5\%, while CORAL, the best baseline, reaches 55.5\%. On \mbox{\textbf{SciLaws-Bench}}, 25$\times$ more samples (Best-of-50 vs.\ Best-of-2) lift the strongest ReAct agent only from 56.9\% to 59.3\%, below \myTool{}'s 62.8\% (Table~\ref{tab:slb}). On \textbf{IdeaBench} (Fig.~\ref{fig:cycle}c; Table~\ref{tab:ib}), from the 27- to the 122-idea setting, \myTool{}'s distinct novel ideas rise from 26.0 to 61.6 with 95 calls per paper, whereas Best-of-$N$'s rise only from 4.7 to 8.8 with 242 calls and MILO's reach 48.2 with 250. Its lead thus widens as the budget grows.

\finding{RQ4: Do the gains depend on the judge?\label{sec:analysis}}\pullup\textls[-13]{We re-score \textbf{IdeaBench} with two judges unused in search: \myTool{} still finds the most distinct novel ideas (17.5 and 13.5 vs.\ 13.1 and 11.3 for OpenEvolve; paired 90\% intervals above zero; Appendix~\ref{app:results:ib}).}\stoppullup

\section{Conclusion}
\label{sec:conclusion}

\pullup We presented \myTool{}, a framework for scientific discovery in which autonomous research units adapt their own mutators and share only verified findings. On five benchmarks spanning multiple research stages, it outperforms frontier AI agents, evolutionary search, and AI co-scientists, discovering more novel ideas and scientific laws that satisfy domain constraints. By self-adapting its search, \myTool{} benefits more from added budget, whereas existing frameworks show diminishing returns.\stoppullup

\section*{Acknowledgements}
We thank Prof.~Vijay Ganesh (Georgia Institute of Technology) for his fruitful feedback and insightful discussions throughout this project, which helped us improve this paper.

\clearpage
\begingroup
\makeatletter
\let\inv@url\url
\def\inv@spliturl#1://#2\inv@end{\mbox{\nolinkurl{#1://}}\nobreak\nolinkurl{#2}}
\renewcommand{\url}[1]{\in@{://}{#1}\ifin@\href{#1}{\inv@spliturl#1\inv@end}\else\inv@url{#1}\fi}
\makeatother
\bibliographystyle{preprint_authoryear}
\bibliography{references}
\endgroup

\clearpage

\appendix
\renewcommand{\numsz}{\fontsize{10}{10}\selectfont}
\newcommand{\AppTocS}[1]{\noindent\makebox[2.2em][l]{\ref*{#1}}\hyperref[#1]{\nameref*{#1}}\ \dotfill\ \pageref*{#1}\par}
\newcommand{\AppTocSS}[1]{\noindent\hspace*{1.6em}\makebox[2.4em][l]{\ref*{#1}}\hyperref[#1]{\nameref*{#1}}\ \dotfill\ \pageref*{#1}\par}
\section*{Appendix}
{\hypersetup{linkcolor=black}%
\begin{center}\begin{minipage}{0.86\textwidth}
\AppTocS{app:setup}
\AppTocSS{app:setup:repro}
\AppTocSS{app:setup:benchmarks}
\AppTocSS{app:examples}
\AppTocSS{app:metrics}
\AppTocSS{app:rinobench}
\AppTocS{app:results}
\AppTocSS{app:results:eb}
\AppTocSS{app:results:sc}
\AppTocSS{app:results:ib}
\AppTocSS{app:results:is}
\AppTocS{app:prompts}
\end{minipage}\end{center}}
\vspace{4mm}

\section{Experimental Setup}
\label{app:setup}\nobreak

\subsection{Reproducibility and Implementation Details}
\label{app:setup:repro}

\noindent\textbf{Machine configuration.}
Our experiments run on CPU-only x86-64 instances. A typical instance has 192 vCPUs (96 physical cores across two sockets, Intel Xeon Platinum 8488C Sapphire Rapids, two threads per core), 1.5\,TiB of memory across two NUMA nodes, and a 1\,TB SSD-backed root volume, and runs Linux (kernel 6.1). The instances host only the search loop, verifiers, and task sandboxes, and all model inference except the IdeaScientist judge (Appendix~\ref{app:ideascientist}) runs remotely through a managed inference service. The paragraphs below give the mutator pool and hyperparameters, then describe scheduling, concurrent updates to shared knowledge, sandboxing and data isolation, and verification thresholds.

\vspace{2mm}
\noindent\textbf{Mutator pool and hyperparameters.}
The initial harness set $\mathcal{H}$ contains a minimal tool-calling harness of our own, mini-SWE-agent, OpenCode, OpenHands, and Goose. The model set $\mathcal{M}$ contains Kimi K3, Grok 4.7, GLM 5.3, MiniMax M2.5, and four other frontier models, each run at its highest reasoning effort and full output limit, and $\mathcal{G}$ holds their compatible pairs. On EurekaBench, \myTool{} runs $N=20$ research units per task for up to $B=80$ rounds. ScholarCatalyst uses $N=10$ units and up to 80 rounds, and SciLaws-Bench $N=10$ units and up to 50 rounds. All three use patience $P=2$ and lag $\lambda=3$ and cap each agent session at four hours (90 minutes on SciLaws-Bench). A harness trial twins every attempt and is decided after four twin pairs. On IdeaBench, $N=4$ units evolve an idea-generation program over $B=12$ rounds, and the frozen program then runs on each test paper. The 122- and 241-idea settings use 10 and 8 rounds. On IdeaScientist, $N=4$ units run $B=6$ rounds per problem.

\vspace{2mm}
\noindent\textbf{Asynchronous execution.}\label{app:loop}
\looseness=-1 Algorithm~\ref{alg:inventa} runs $N$ research units in parallel. In each round, a unit reads a snapshot of $\mathcal{K}$, selects a parent, and generates a child. Verifiers evaluate the child, return its certificate, and publish qualifying findings and model-level statistics. Like symbolic feedback for fine-tuning LLMs~\citep{jha2025rlsf}, the certificate reports every check rather than a single score. The unit then updates its lineage, and orchestrator actions adapt the search (\S\ref{sec:units}). Units start and finish rounds independently, and a scheduler decides when each may start its next round.

\vspace{2mm}
\noindent\textbf{Bounded staleness.}
Unit $j$ starts a round only if
\begin{equation}
r_j<\min_{j'\in\mathrm{active}}r_{j'}+\lambda,
\label{eq:lag}
\end{equation}
where active units exclude those paused after repeated failures or treated as hung. Among ready units, those with fewer completed rounds take priority, and twin rounds for harness comparison are exempt from the bound. With $\lambda=1$, units advance roughly in lockstep. Larger values let faster units advance further while limiting differences in the shared evidence each unit sees. At fixed evaluation checkpoints, non-hung units synchronize and the selected candidate is frozen for evaluation. Test data never feeds into search or selection.

\pullup \vspace{2mm}
\noindent\textbf{Optimistic concurrency.}
Each unit works from the shared-knowledge snapshot taken when its round begins. Before publishing a candidate, verifiers atomically re-read the best verified candidate and apply the publication rule. If another unit has since improved that reference, the candidate must exceed the updated reference by $\delta$ to enter $\mathcal{K}_{\mathrm{ver}}$, while its other qualifying findings remain eligible. The best verified score is thus nondecreasing despite stale snapshots~\citep{kung1981optimistic}.\stoppullup

\vspace{2mm}
\noindent\textbf{Sandboxing and data isolation.}
Each mutator runs in a sandbox without internet access that exposes only the training data and the task workspace. Validation data and verifier code stay outside the sandbox, and units see validation results only through certificates. The test split is held outside the search environment, where no unit or verifier can access it, and is scored only on candidates frozen at checkpoints. An attempt that tries to read restricted data or verifier files is voided and updates neither its lineage nor the shared knowledge.

\pullup \vspace{2mm}
\noindent\textbf{Verification thresholds.}\label{app:verify}
Pairing LLM mutators with programmatic verifiers makes \myTool{} a neurosymbolic design~\citep{jana2024neurosymbolic,jana2026proofbridge}. For component ablations, the implementation uses a separate margin $\delta_{\mathrm{abl}}$ in place of $\delta$, with $\delta_{\mathrm{abl}}=2\delta$ on EurekaBench. It also rejects verification when a component's measured contribution exceeds $\gamma=5$ times the larger of the candidate's gain~and~the~ablation~margin:
\begin{equation}
\DropM_t(x)>\gamma\max\{\Gain(x),\delta_{\mathrm{abl}}\}.
\end{equation}
This heuristic is intended to flag candidates whose components may offset one another's effects.\stoppullup

\subsection{Benchmark Details}
\label{app:setup:benchmarks}

Below, we detail each benchmark of \S\ref{sec:experiments}, including our IdeaBench and IdeaScientist protocols.

\vspace{2mm}
\noindent\textbf{EurekaBench.}
\looseness=-1 We use the 16 public CPU-only tasks of EurekaBench~\citep{geng2026eurekabench}: three plasma-physics tasks on the ARC and negative-triangularity tokamaks, seven astrophysics tasks on CAMELS and DREAMS cosmological simulations, and six neuroscience tasks on \emph{C.~elegans} models. Each task runs in an experimental environment, and an agent must propose a mechanism that predicts held-out data, answers expert insight questions, and respects domain constraints (Table~\ref{tab:eb}).

\vspace{2mm}
\noindent\textbf{SciLaws-Bench.}\label{app:slb-subset}
SciLaws-Bench~\citep{huang2026scilaws} \looseness=-1 contains 118 scientific law-discovery problems from six disciplines, including materials and engineering, ecology and hydrology, and the social sciences, drawn from 381 papers and about 8 million measured, observed, and simulated data points. A system must find a closed-form law that generalizes: single-group problems test extrapolation or transfer across time and conditions, and multi-group problems hold out entire groups, such as unseen crystals in an optical-dispersion task. We use 38 of the 118 tasks, selected by rules fixed before our search: from the 99 permissively licensed tasks, we keep those meeting four criteria:
\begin{enumerate}[nosep,leftmargin=1.4em]
\item A validation split can be built from the training data the way the test split was.
\item The task provides sufficient data: at least 50 test and 50 validation rows for single-group tasks, or at least eight test groups and 15 training groups for multi-group tasks.
\item Its estimated ceiling exceeds 55. The ceiling is the maximum of a noise estimate, a flexible non-law fit, and the best per-task score of the benchmark's nine baseline models.
\item The published reference law can be reproduced exactly in the scored submission format.
\end{enumerate}
The first three criteria leave 40 tasks, and the reproducibility criterion removes two more, leaving 16 single-group and 22 multi-group tasks. Every method in Table~\ref{tab:slb} uses this 38-task subset.

\vspace{2mm}
\noindent\textbf{ScholarCatalyst.}
ScholarCatalyst~\citep{kim2026scholarcatalyst} asks a system to retrieve the papers that inspired, or could have inspired, a research project. Its 207 source papers are 2025--2026 computer-science papers spanning machine learning, vision, language, AI, and robotics, over 100 of them orals, spotlights, or award winners, and their lead authors labelled the positives. Systems search a 190{,}896-paper corpus, restricted per query to papers that predate its source paper. Inspiration often crosses fields: nearly half of the credited papers lie outside the project's subject area, and positives are no more similar to the query than related papers the authors rejected. We evaluate on all 894 test queries: 207 core research queries and 687 subfield-specific ones. Following \citet{kim2026scholarcatalyst}, who recommend a model whose knowledge cutoff precedes every paper in the task and corpus, all methods use DeepSeek V3.2, so all rows share one model.

\vspace{2mm}
\noindent\textbf{IdeaBench.}\label{app:ideabench}
IdeaBench~\citep{guo2024ideabench} pairs 2{,}374 influential biomedical papers published in 2024 with 23{,}460 references, research articles with at least five citations that each paper cites in its background section. A method reads the abstracts of five of a paper's references and proposes research ideas, which are judged against the paper's own, unseen idea. We split the papers by a salted hash of their identifiers into 904 training, 734 validation, and 736 test papers. \myTool{}'s units evolve an idea-generation program using five training papers and are scored on 8--12 validation papers, while validation papers 50--149 form a 100-paper check set excluded from search. The frozen program then runs on every test paper.

\begingroup\doublehyphendemerits=900000\finalhyphendemerits=900000
\vspace{2mm}
\noindent\textbf{IdeaScientist.}\label{app:ideascientist}
IdeaScientist~\citep{liu2026ideascientist} asks a system to write a grounded research proposal for a research problem, given as a problem definition and its central challenge, drawn from a machine-learning methodology paper published after January 1, 2026. Systems search an offline corpus of 2.77 million research ideas decomposed from arXiv papers in all subject areas, reading only papers that predate the problem's paper. Useful mechanisms may come from seemingly unrelated fields that share the problem's structure, and the system must anticipate a direction before it appeared. We evaluate on its 277 test problems, with one \myTool{} search per problem. A benchmark plug-in provides IdeaScientist's judge, paper corpus, and renderer, and scoring uses the benchmark's own code. A candidate is a proposal in the benchmark's results-masked schema, with a rationale and a signed ledger of papers read. We serve the benchmark's judge, Qwen3.6-27B, on GPU nodes with SGLang, with reasoning disabled, temperature 0.7, and top-p 0.8. Re-scoring stored proposals from earlier systems reproduces scores within 1\%. Agents run unprivileged inside the judge's container, with private benchmark files hidden and external literature and~search~hosts~blocked.\par\endgroup

\subsection{Task Examples}
\label{app:examples}

Each example instantiates the task tuple $\tau=\langle q,\mathcal{W},\mathcal{C},\mathcal{V}\rangle$ of Def.~\ref{def:tasks} for one task class.

\pullup \vspace{2mm}
\noindent\textbf{Scientific law discovery.}
On EurekaBench's ice-shelf task, $q$ asks why a textbook flow law misfits measured ice-shelf flow; $\mathcal{W}$ provides velocity and thickness fields from a FEniCSx finite-element simulator; $\mathcal{C}$ holds expert-written physical constraints, such as thermodynamic laws; and $\mathcal{V}$ runs the candidate on held-out flow regimes, tests the constraints, and poses insight questions that LLM agents may answer by querying the simulator. The candidate is the flow law, coded and explained, so verifiers execute it, as compilers and symbolic execution~\citep{jana2024cotran} check generated code.\stoppullup

\pullup \vspace{2mm}
\noindent\textbf{Literature retrieval.}
On ScholarCatalyst, $q$ asks which prior work could advance a recent project's central research question, or the same question seen from one subfield; $\mathcal{W}$ is a 190{,}896-paper corpus restricted to papers that predate the source paper; and $\mathcal{V}$ scores the ranked list against the papers the source paper's authors labelled as inspiring. The candidate is a retrieval method, as a program.\stoppullup

\vspace{2mm}
\noindent\textbf{Idea generation.}
On IdeaBench, $q$ asks for research ideas that build on a paper's references; $\mathcal{W}$ holds the abstracts of five of them; and $\mathcal{V}$ rates every idea with an expert-calibrated LLM judge and merges near-duplicates for counting. The candidate is an idea-generation program, evolved on training and validation papers, then frozen for testing.

\vspace{2mm}
\noindent\textbf{Grounded proposals.}
On IdeaScientist, $q$ \looseness=-1 is a research problem, a problem definition with its central challenge, from a methodology paper published in 2026; $\mathcal{W}$ offers search, bibliography, and paper-reading tools over pre-cutoff papers; $\mathcal{C}$ requires a schema-valid proposal that cites at least one paper, with at least 90\% of cited papers read per its ledger; and $\mathcal{V}$ is the benchmark's LLM judge of novelty and quality. The candidate is a proposal with a rationale and a ledger of papers read.

\subsection{Evaluation Metrics}
\label{app:metrics}

This section expands the metric summary of \S\ref{sec:experiments:metrics} with each benchmark's definitions and scoring.

\vspace{2mm}
\noindent\textbf{EurekaBench.}
\looseness=-1 Each task provides expert-written criteria of three kinds~\citep{geng2026eurekabench}. Predictive accuracy (PA) executes the submitted mechanism on held-out test data, from the original studies or the task's simulator, and normalizes each criterion to $[0,1]$ under the task's rubric. Scientific-constraint (SC) tests check that the mechanism does not contradict the domain's assumptions, and scientific-insight (SI) questions credit an insight only when the judge finds complete evidence for it. Two agentic LLM judges evaluate both and may run extra simulator experiments for SI. The conditioned score (CS) pools a judge's PA criterion scores and SI question scores into one mean, is zero unless every SC test passes, and is averaged over the two judges. We report CS as the primary measure, PA and SI unconditionally, and SC as the number of tasks that fail either judge's tests.

\vspace{2mm}
\noindent\textbf{SciLaws-Bench.}
Numeric fit ($S_N$) \looseness=-1 is computed programmatically on the held-out test split with a task-specific fit metric (RMSE, sMAPE, log-MAE, or $R^2$), normalized so that the best published law scores 0.5 and a perfect predictor 1~\citep{huang2026scilaws}; we report it $\times100$. Multi-group tasks average over groups, and a failed group scores zero. We report single-group, multi-group, and all tasks, with a 95\% task-bootstrap confidence interval for the overall mean, and, following the benchmark, count a task as exceeding the published reference when $S_N>52$ (0.52 on the benchmark's scale). Scientific validity ($S_V$) is the fraction of a task's rubric items satisfied, checked by a code-executing LLM judge through probes and source inspection. A submission that fails the anti-hacking item, which rejects degrees of freedom moved outside the declared law, receives zero validity.

\begin{table*}[t]
\centering
\captionsetup{font=small,skip=2pt}
\renewcommand{\arraystretch}{0.90}
\caption{\textbf{ScholarCatalyst by query type.} \textit{Mean}: average of the eight cells, the primary metric of Table~\ref{tab:sc}. \textit{+~real metadata}: real document IDs and years. \best{Bold}/\nd{underline}: best/second.}
\label{tab:sc-split}
\footnotesize
\setlength{\aboverulesep}{0.6pt}\setlength{\belowrulesep}{1pt}
\fitat{\asc}{18}{%
\begin{tabular}{@{}l !{\color{black!22}\vrule} N N N N !{\color{black!22}\vrule} N N N N !{\color{black!22}\vrule} N @{}}
\toprule
 & \multicolumn{4}{c!{\color{black!22}\vrule}}{\mh{Core queries (207; \%)$\uparrow$}} & \multicolumn{4}{c!{\color{black!22}\vrule}}{\mh{Subfield queries (687; \%)$\uparrow$}} & \\
\cmidrule(lr){2-5}\cmidrule(lr){6-9}
\textbf{Method} & \mh{nDCG@20} & \mh{Recall@5} & \mh{Recall@20} & \mh{Recall@100} & \mh{nDCG@20} & \mh{Recall@5} & \mh{Recall@20} & \mh{Recall@100} & \mh{Mean$\uparrow$} \\
\midrule
\grp{10}{Retrievers without a language model}\\
\ind\rid{1}BM25 & 15.6 & 12.0 & 22.6 & 39.4 & 20.8 & 11.9 & 26.0 & 45.4 & 24.2 \\
\ind\rid{2}Cohere Embed v4 & 24.9 & 19.0 & 34.3 & 52.3 & 31.7 & 19.2 & 38.7 & 63.4 & 35.4 \\
\ind\rid{3}Cohere v4 + BM25 fusion & 22.4 & 17.2 & 31.5 & 52.6 & 30.1 & 18.5 & 37.0 & 63.4 & 34.1 \\
\ind\rid{4}Cohere v4 + Rerank 3.5 & 23.3 & 20.6 & 35.2 & 52.3 & 31.4 & 18.4 & \nd{40.1} & 63.4 & 35.6 \\
\grp{10}{DeepSeek V3.2 agents (Cohere v4 retriever)}\\
\ind\rid{5}Tool-calling Agent & 23.6 & 16.3 & 33.8 & 52.9 & 30.2 & 19.6 & 36.7 & 63.1 & 34.5 \\
\ind\rid{6}+ real metadata & 22.8 & 17.9 & 34.3 & \nd{53.3} & 30.1 & 19.2 & 36.7 & 63.1 & 34.7 \\
\ind\rid{7}Deep-research Agent & 25.0 & 19.4 & 32.5 & 51.8 & 31.7 & 20.2 & 37.4 & 63.8 & 35.2 \\
\ind\rid{8}OpenCode & \nd{25.7} & 19.6 & 33.5 & \nd{53.3} & \nd{33.4} & \nd{22.3} & 39.8 & \nd{64.2} & \nd{36.5} \\
\ind\rid{9}OpenHands & 25.2 & \nd{21.9} & \nd{36.2} & 52.8 & 32.4 & 21.3 & 38.9 & 61.9 & 36.3 \\
\ind\rid{10}Seed agent (initialization) & 22.5 & 18.1 & 31.9 & 50.2 & 28.7 & 19.6 & 33.4 & 60.9 & 33.1 \\
\rowcolor{tbOurs}\oursL{\bandstrut\ind\rid{11}\textbf{\myTool{}}} & \best{32.0} & \best{25.9} & \best{45.3} & \best{61.6} & \best{41.2} & \best{27.2} & \best{50.7} & \best{72.4} & \oursR{\best{44.5}} \\
\bottomrule
\end{tabular}}
\end{table*}
\pullup \vspace{2mm}
\noindent\textbf{ScholarCatalyst.}
The positives for each query are the candidate papers that the source paper's lead author labelled as having advanced, or as able to advance, the project~\citep{kim2026scholarcatalyst}. Recall@$k$ is the fraction of a query's positives retrieved in the top $k$ results, and nDCG@20 additionally rewards ranking them higher; both are computed programmatically from these labels. We report nDCG@20 and Recall@5, @20, and @100 for core research queries and subfield-specific queries. The benchmark defines no aggregate; our \textit{Mean} (Tables~\ref{tab:sc} and~\ref{tab:sc-split}) is the unweighted mean of the eight values.\stoppullup

\vspace{2mm}
\noindent\textbf{IdeaBench.}
The benchmark's \looseness=-1 insight scores for novelty and feasibility measure how often an LLM jury ranks a method's ideas above the paper's own idea, without revealing which is the original~\citep{guo2024ideabench}. In our setup, a three-model jury ranks each method's top three ideas; the published jury uses the writer models, and the disjoint jury shares no model with any writer. Two further measures compare ideas with the paper's abstract, BERTScore F1 (\emph{Semantic}) and an LLM-rated idea overlap (\emph{Overlap}); neither has a preferred direction. We add measures from our own novelty judge, a frontier LLM calibrated on RINoBench (Appendix~\ref{app:rinobench}) that rates every generated idea from 1 to 10: the number of distinct ideas per paper rated at least 7 (our primary measure) or at least 8, merging near-duplicates among them (Cohere Embed v4, cosine 0.8); the best idea's score (maximum novelty); and the share of all ideas surviving a merge over the whole pool (non-duplicate rate).

\vspace{2mm}
\noindent\textbf{IdeaScientist.}
All measures are scored by the authors' judge, Qwen3.6-27B, served locally with their sampling settings~\citep{liu2026ideascientist}. Novelty has three measures: \emph{All}, a 1--10 judgment of how much the proposal adds beyond the closest papers in the corpus; \emph{Cutoff}, the same judgment against pre-cutoff papers; and \emph{In-domain}, the share of proposals judged to transfer a mechanism across problem settings through a shared structure. Proposal quality averages six 1--10 peer-review criteria: relevance, clarity, specificity, actionability, soundness, and impact. All measures use a 0--100 scale, and the combined score is the mean of the novelty and quality averages. A problem without a scorable proposal scores zero throughout.

\begin{table*}[t]
\centering
\begin{minipage}[t]{\wJ}\vspace{0pt}
\centering
\caption{\textbf{IdeaBench under three judges} (test papers). The same ideas re-scored by two judges unused in the loop: a second frontier-model judge (736 papers) and Grok 4.7 (a fixed sample of 200). Outside judges are stricter, so compare within a judge. DN8/DN7: distinct ideas per paper rated $\ge$8/$\ge$7 (DN7: primary). \best{Bold}/\nd{underline}: best/second per judge.}
\label{tab:ib-judges}
\footnotesize\renewcommand{\arraystretch}{\rsJ}\setlength{\aboverulesep}{0.6pt}\setlength{\belowrulesep}{1pt}
\fitat{\asc}{6}{%
\begin{tabular}{@{}l !{\color{black!22}\vrule} N N >{\columncolor{tbHead}[\tabcolsep][0pt]\numsz}c @{}}
\toprule
\textbf{Method} & \mhh{Max}{Novelty$\uparrow$} & \mh{DN8$\uparrow$} & \mh{DN7$\uparrow$} \\
\midrule
\grp{4}{Primary judge}\\
\ind\rid{1}OpenEvolve & \nd{8.63} & \nd{13.5} & \nd{14.7} \\
\ind\rid{2}MILO, 3 islands & 8.44 & 10.1 & 13.4 \\
\rowcolor{tbOurs}\oursL{\bandstrut\ind\rid{3}\textbf{\myTool{}}} & \best{8.88} & \best{18.4} & \oursR{\best{26.0}} \\
\grp{4}{Second judge}\\
\ind\rid{1}OpenEvolve & \nd{8.23} & \best{7.8} & \nd{13.1} \\
\ind\rid{2}MILO, 3 islands & 8.01 & 6.1 & 12.1 \\
\rowcolor{tbOurs}\oursL{\bandstrut\ind\rid{3}\textbf{\myTool{}}} & \best{8.26} & \nd{7.1} & \oursR{\best{17.5}} \\
\grp{4}{Judge: Grok 4.7}\\
\ind\rid{1}OpenEvolve & \best{7.98} & \best{5.6} & \nd{11.3} \\
\ind\rid{2}MILO, 3 islands & 7.85 & 2.5 & 8.8 \\
\rowcolor{tbOurs}\oursL{\bandstrut\ind\rid{3}\textbf{\myTool{}}} & \nd{7.93} & \nd{4.8} & \oursR{\best{13.5}} \\
\bottomrule
\end{tabular}}
\end{minipage}\hfill
\begin{minipage}[t]{\wA}\vspace{0pt}
\centering
\caption{\textbf{Ablation of \myTool{}'s refinement} (100 validation papers, no test data). Each row replaces one of four design choices with a simpler alternative. Weak: share of ideas rated $\le$6. \best{Bold}/\nd{underline}: best/second.}
\label{tab:ib-abl}
\footnotesize\renewcommand{\arraystretch}{0.86}\setlength{\aboverulesep}{0.6pt}\setlength{\belowrulesep}{1pt}
\fitat{\asc}{10}{%
\begin{tabular}{@{}l !{\color{black!22}\vrule} N N >{\columncolor{tbHead}\numsz}c N N @{}}
\toprule
\textbf{Variant} & \mhh{Max}{Novelty$\uparrow$} & \mh{DN8$\uparrow$} & \mh{DN7$\uparrow$} & \mhh{Non-}{dup$\uparrow$} & \mh{Weak$\downarrow$} \\
\midrule
\grp{6}{Refinement without one design choice}\\
\ind\rid{1}without the novelty rewrite prompt & 7.99 & \nd{15.0} & \nd{24.8} & 95.5 & \nd{7.8} \\
\ind\rid{2}without the paper-level keep rule & 8.18 & 12.8 & 23.9 & 97.1 & 11.5 \\
\ind\rid{3}rewriting only the best ideas & 8.67 & 12.8 & 23.1 & 96.3 & 14.3 \\
\ind\rid{4}10 rewrites instead of 18 & \nd{8.72} & 14.7 & \nd{24.8} & \nd{97.2} & 8.0 \\
\midrule
\rowcolor{tbOurs}\oursL{\bandstrut\ind\rid{5}\textbf{\myTool{}} (all four choices)} & \best{8.81} & \best{17.5} & \best{25.0} & \best{97.9} & \oursR{\best{7.4}} \\
\bottomrule
\end{tabular}}
\par\vspace{\gapC}
\caption{\textbf{Novelty judge calibration on RINoBench.} Cut points fitted on training ideas map the judge's 1--10 scores to the experts' 1--5 levels. The best judges of \citet{schopf2026rinobench} reach macro F1 0.172 and 0.162 on test. \best{Bold}: best per round; shaded: ours.}
\label{tab:judge-calib}
\footnotesize\renewcommand{\arraystretch}{0.86}\setlength{\aboverulesep}{0.6pt}\setlength{\belowrulesep}{1pt}
\fitat{\asc}{6}{%
\begin{tabular}{@{}l !{\color{black!22}\vrule} N >{\columncolor{tbHead}\numsz}c N @{}}
\toprule
\textbf{Judge configuration} & \mh{Spearman$\uparrow$} & \mh{Macro F1$\uparrow$} & \mh{Mean absolute error$\downarrow$} \\
\midrule
\grp{4}{Round 1: train, 5-fold cross-validation}\\
\rowcolor{tbOurs}\oursL{\bandstrut\ind\rid{1}\textbf{P0 (ours)}} & \best{0.357} & \best{0.306} & \oursR{\best{0.84}} \\
\ind\rid{2}P1 & 0.353 & 0.272 & 0.90 \\
\ind\rid{3}P2 & 0.331 & 0.266 & 0.88 \\
\ind\rid{4}P1, two-judge panel & 0.156 & 0.231 & 1.06 \\
\grp{4}{Round 2: test}\\
\rowcolor{tbOurs}\oursL{\bandstrut\ind\rid{5}\textbf{Ours, calibrated}} & 0.275 & \best{0.262} & \oursR{0.88} \\
\ind\rid{6}Ours, uncalibrated & 0.208 & 0.195 & 0.84 \\
\bottomrule
\end{tabular}}
\end{minipage}
\end{table*}

\subsection{Novelty Judge Calibration on RINoBench}
\label{app:rinobench}

RINoBench~\citep{schopf2026rinobench} pairs 1{,}381 research ideas from ICLR 2022 and 2023 submissions with novelty levels on a 1--5 scale that bin the averaged novelty scores of expert reviewers. We use it only to select and calibrate the IdeaBench novelty judge. We screen five candidate judges and three rubrics on 100 of its 1{,}104 training ideas, evaluate the surviving configurations on all training ideas, and fit cut points from the judges' 1--10 scores to the five levels by cross-validation (Table~\ref{tab:judge-calib}). Under the fitted mapping, a score of 7 corresponds to \emph{novel} and a score of at least 8 to \emph{highly innovative}. On the 277 test ideas, the selected judge reaches a macro F1 of 0.262, against 0.172 for the best judge the benchmark's authors report. These categories are calibrated to reviewer labels, not to reviews of IdeaBench outputs.

\raggedbottom
\makeatletter\setlength{\@fptop}{0pt}\setlength{\@fpsep}{8pt}\setlength{\@fpbot}{0pt plus 1fil}\makeatother
\setlength{\floatsep}{14pt plus 2pt minus 2pt}\setlength{\textfloatsep}{16pt plus 2pt minus 2pt}
\renewcommand{\topfraction}{0.9}\renewcommand{\textfraction}{0.07}

\setlength{\intextsep}{12pt plus 2pt minus 2pt}

\Needspace*{8\baselineskip}
\section{Additional Results}
\label{app:results}\nobreak
\subsection{Scientific Law Discovery: EurekaBench}
\label{app:results:eb}
After 10 rounds, \myTool{}'s best validation CS exceeds the best validation score over all runs of Table~\ref{tab:eb}'s 14 frontier agents on 15 of 16 tasks, and after 20 rounds on all 16 (means of 64.7\% and 72.5\% vs.\ 48.4\%). The verifiers published 2,502 findings (239 \textsc{Verified}, 395 \textsc{Partial}, 1,868 \textsc{Refuted}). On 11 of 16 tasks, the validation-best law comes from a lineage that \textsc{Restart} re-seeded, and 90 of the 320 units end on an evolved harness.

\subsection{Retrieving Prior Work: ScholarCatalyst by Query Type}
\label{app:results:sc}
\myTool{} leads in all eight cells (Table~\ref{tab:sc-split}), most on subfield Recall@20 (50.7\% vs.\ 40.1\% for Cohere Embed v4 + Rerank 3.5, the next best).

\begin{table*}[t]
\centering
\begin{minipage}[t]{\wW}\vspace{0pt}
\centering
\caption{\textbf{IdeaBench with weaker writers} (736 test papers). Every method uses the same writer; judge, juries, and scoring as in Table~\ref{tab:ib}. DN7: primary. Insight: disjoint jury. \best{Bold}/\nd{underline}: best/second per writer.}
\label{tab:ib-weak}
\footnotesize\renewcommand{\arraystretch}{0.86}\setlength{\aboverulesep}{0.6pt}\setlength{\belowrulesep}{1pt}
\fitat{\asc}{12}{%
\begin{tabular}{@{}l !{\color{black!22}\vrule} N N >{\columncolor{tbHead}\numsz}c N N !{\color{black!22}\vrule} N @{}}
\toprule
\textbf{Method} & \mhh{Max}{Novelty$\uparrow$} & \mh{DN8$\uparrow$} & \mh{DN7$\uparrow$} & \mhh{Non-}{dup$\uparrow$} & \mhh{Insight}{Novelty$\uparrow$} & \mhh{Calls}{/paper$\downarrow$} \\
\midrule
\grp{7}{Writer: DeepSeek V3.2}\\
\ind\rid{1}Single prompt & 6.75 & 0.1 & 1.3 & \nd{92.0} & 93.6 & \best{3} \\
\ind\rid{2}Best-of-26 & 7.25 & 0.8 & 8.6 & 69.2 & 92.5 & 52 \\
\ind\rid{3}GEPA & 7.91 & 5.3 & \best{22.7} & 81.2 & \nd{98.1} & 56 \\
\ind\rid{4}OpenEvolve & \nd{8.09} & \best{11.7} & \nd{18.3} & 91.6 & \best{99.6} & 56 \\
\ind\rid{5}MILO, 3 islands & 7.92 & 4.2 & 12.8 & 62.9 & 95.8 & 56 \\
\rowcolor{tbOurs}\oursL{\bandstrut\ind\rid{6}\textbf{\myTool{}}} & \best{8.48} & \nd{9.1} & 18.1 & \best{98.8} & 97.5 & \oursR{\nd{46}} \\
\grp{7}{Writer: MiniMax M2.5}\\
\ind\rid{1}Single prompt & 3.27 & 0.0 & 0.2 & \nd{96.3} & 50.9 & \best{3} \\
\ind\rid{2}Best-of-26 & 6.23 & 0.1 & 0.9 & 74.8 & 80.8 & \nd{33} \\
\ind\rid{3}GEPA & 7.21 & 0.4 & \best{6.7} & 73.8 & \best{93.7} & 56 \\
\ind\rid{4}OpenEvolve & \nd{7.25} & \nd{0.6} & 3.5 & 46.8 & \nd{89.9} & 85 \\
\ind\rid{5}MILO, 3 islands & 6.66 & 0.2 & 1.6 & 69.7 & 80.1 & 36 \\
\rowcolor{tbOurs}\oursL{\bandstrut\ind\rid{6}\textbf{\myTool{}}} & \best{7.43} & \best{1.0} & \nd{5.9} & \best{97.7} & 88.1 & \oursR{45} \\
\grp{7}{Writer: Kimi K3}\\
\ind\rid{1}Single prompt & 7.26 & 0.5 & 1.8 & 75.7 & 93.3 & \best{3} \\
\ind\rid{2}Best-of-26 & 7.67 & 2.8 & 8.0 & 37.4 & 92.8 & 52 \\
\ind\rid{3}GEPA & 8.00 & 11.6 & \nd{21.1} & 62.5 & 97.0 & 56 \\
\ind\rid{4}OpenEvolve & \nd{8.40} & \best{17.2} & 18.9 & \nd{90.9} & \best{99.7} & 56 \\
\ind\rid{5}MILO, 3 islands & 8.07 & 9.9 & 15.6 & 62.4 & 97.9 & 56 \\
\rowcolor{tbOurs}\oursL{\bandstrut\ind\rid{6}\textbf{\myTool{}}} & \best{8.91} & \nd{15.6} & \best{25.1} & \best{99.4} & \nd{98.0} & \oursR{\nd{46}} \\
\bottomrule
\end{tabular}}
\end{minipage}\hfill
\begin{minipage}[t]{\wR}\vspace{0pt}
\centering
\caption{\textbf{Robustness of Insight Novelty} ($\times100$). Pub.: IdeaBench's protocol, with the writer pool as jury. A: each top-3 idea and the paper's idea rewritten into one neutral 120--160-word paragraph (style normalization; \citealp{si2025can}). B: a jury sharing no model with any writer. C: top three by novelty alone. D: A+B+C. At 27 ideas, \myTool{} scores 96.8 (Pub.) and 97.4 (B).}
\label{tab:insight-robust}
\footnotesize\renewcommand{\arraystretch}{0.86}\setlength{\aboverulesep}{0.6pt}\setlength{\belowrulesep}{1pt}
\fitat{\asc}{10}{%
\begin{tabular}{@{}l !{\color{black!22}\vrule} N !{\color{black!22}\vrule} N N N N @{}}
\toprule
\textbf{Method} & \mh{Pub.} & \mh{A} & \mh{B} & \mh{C} & \mh{D} \\
\midrule
\grp{6}{Baselines}\\
\ind\rid{1}Single prompt & 94.8 & 95.9 & 91.6 & 94.9 & 91.6 \\
\ind\rid{2}Over-gen.\ + rerank & 97.0 & 97.2 & 95.2 & 97.0 & 95.3 \\
\ind\rid{3}Best-of-26 & 96.5 & 96.4 & 94.1 & 96.9 & 94.4 \\
\ind\rid{4}Best-of-122 & 94.9 & 95.5 & 92.7 & 96.7 & 94.4 \\
\ind\rid{5}GEPA & 97.1 & 96.5 & 96.5 & 97.4 & 96.2 \\
\ind\rid{6}OpenEvolve & 98.7 & 98.5 & 98.6 & 99.1 & 98.8 \\
\ind\rid{7}EvoX & 97.0 & 96.0 & 94.6 & 96.8 & 94.2 \\
\ind\rid{8}CORAL & 98.4 & 96.5 & 96.4 & 98.3 & 93.3 \\
\ind\rid{9}ResearchAgent & 95.6 & 95.3 & 95.2 & 98.5 & 98.2 \\
\ind\rid{10}MILO, 3 islands & 98.3 & 98.2 & 98.1 & 98.9 & 98.5 \\
\ind\rid{11}MILO, 15 islands & 98.5 & 98.6 & 98.4 & 99.6 & 98.9 \\
\ind\rid{12}MILO, 30 islands & 97.9 & 98.3 & 97.9 & 99.6 & 98.8 \\
\midrule
\grpours{6}{\myTool{}}\\
\ind\rid{13}122 ideas/paper & 96.0 & 96.4 & 97.1 & 98.1 & 98.5 \\
\ind\rid{14}241 ideas/paper & 94.8 & 95.2 & 96.1 & 97.5 & 98.2 \\
\bottomrule
\end{tabular}}
\end{minipage}
\end{table*}

\subsection{Generating Ideas: IdeaBench}
\label{app:results:ib}

\noindent\textbf{Ideas and model calls.}
Without per-paper refinement, \myTool{} returns 23.4 distinct novel ideas with 24 calls. Refinement raises its maximum novelty from 7.95 to 8.88 and its ideas rated $\geq$8 from 11.2 to 18.4. At 241 ideas, MILO with 30 islands has more ideas rated $\geq$8 (59.2 vs.\ 52.0) and the best maximum novelty (8.93), and OpenEvolve has the highest Insight Novelty (98.7\% vs.\ 96.8\%; Table~\ref{tab:insight-robust} re-scores it under four protocol changes).

\vspace{2mm}
\noindent\textbf{Robustness to the judge.}
The primary judge also guides selection inside MILO, OpenEvolve, and \myTool{}'s refinement, so we re-score the test ideas with two judges unused in the loop: a second frontier-model judge on all 736 papers and Grok 4.7 on a fixed sample of 200 (Table~\ref{tab:ib-judges}). \myTool{} still returns the most distinct novel ideas, 17.5 and 13.5 against at most 13.1 and 11.3 for OpenEvolve, with all paired 90\% bootstrap intervals above zero. Its maximum novelty is close to OpenEvolve's, which has slightly more ideas rated $\geq$8 under both judges.

\pullup \vspace{2mm}
\noindent\textbf{Ablation of the refinement stage.}
Removing one design choice of the refinement at a time on 100 validation papers shows that each helps (Table~\ref{tab:ib-abl}): asking for both novelty and feasibility lowers maximum novelty from 8.81 to 7.99; accepting rewrites by their own score rather than the paper-level aggregate lowers highly innovative ideas from 17.5 to 12.8; rewriting only the best ideas also gives 12.8 and doubles the weak-idea share (7.4\% to 14.3\%); and 10 instead of 18 rewrites gives 14.7.\stoppullup

\vspace{2mm}
\noindent\textbf{Single writers.}
With each of three lower-cost writers alone (Table~\ref{tab:ib-weak}), \myTool{} has the highest maximum novelty (8.48 with DeepSeek V3.2, 7.43 with MiniMax M2.5, 8.91 with Kimi K3) and the most distinct novel ideas with Kimi K3 (25.1 vs.\ 21.1 for GEPA); GEPA leads with DeepSeek V3.2 (22.7 vs.\ 18.1) and MiniMax M2.5 (6.7 vs.\ 5.9).

\setcounter{topnumber}{1}
\subsection{Grounded Proposals: IdeaScientist}
\label{app:results:is}

\myTool{} reaches a combined score of 82.7\%, against 77.7\% for IdeaScientist-Base and 74.6\% for IdeaScientist-Trained, as reported by its authors (Table~\ref{tab:is}). The gain comes from In-domain: 93.9\% of its proposals transfer a mechanism across problem settings, against 54.9\% and 70.9\%.

\section{Prompts and Harnesses}
\label{app:prompts}
The boxes below give the mutator's initial harness and the orchestrator's actions and prompts on EurekaBench, abridged for readability; \ph{field} marks a field filled at run time. Before its twin trial, every \textsc{Retool} edit passes a validator that rejects any scientific content, a policy check in the spirit of policy-guided verifier feedback~\citep{jana2026terraformer}.

\begin{tcolorbox}[pbox=ridU, title={Mutator $\mu_j$: initial harness\hfill\normalfont\itshape evolves by {\upshape\textsc{Retool}}}, title after break={Mutator $\mu_j$: initial harness\hfill\normalfont\itshape continued}]
\pnote{Every unit starts from this minimal tool-calling harness: a control-flow program (\fn{run.py}) with its prompts, procedure, tools, and limits. {\upshape\textsc{Retool}} keeps rewriting all of them, the program included, so each unit's harness evolves over time.}
\plab{ridU}{SYSTEM PROMPT}
\looseness=-1 You are a careful research scientist working inside an automated discovery system. You write mechanisms as runnable code plus a plain explanation. You never guess at numbers you did not compute, you say what you assume, and you prefer one well-reasoned change to many unmotivated ones. You do not try to game the evaluation: a mechanism must explain the observation, not merely fit it.

Your method: state precisely where the classical law fails and where it holds; name the assumptions it rests on; relax the one your research direction questions, and only that one; derive the modified mechanism so that every new term has a physical reading and the right dimensions; predict a signature that separates it from the classical law; check the limiting case in which the new effect vanishes and the classical law must return; check signs, positivity, and bounds of every field; name your additive terms so each can be tested alone; say what would refute you. Do not look up the published answer; a copied answer is not a discovery, and a term you cannot explain is a term the ablation test will expose.

\looseness=-1 \pullup You work in a workspace through the tools \fn{bash}, \fn{write\_file}, \fn{read\_file}, and \fn{list\_files}. Start by reading \fn{AGENTS.md}. Finish by writing the deliverables and \fn{REPLY.md} into \fn{./mechanisms}, then reply with one line saying you are done; do not paste the files in the reply.\stoppullup
\plab{ridU}{FIRST MESSAGE}
Read \fn{AGENTS.md} and do what it asks. Work until \fn{mechanisms/} holds every deliverable and \fn{REPLY.md}.
\plab{ridU}{TASK FILE (\fn{AGENTS.md})}
You are one research unit in a population searching for the mechanism behind a scientific observation. Work in this directory. Write your deliverables (\ph{deliverables}) into \fn{./mechanisms/}; the executable deliverable \ph{code file} must follow the contract in \fn{TASK.md} exactly. You may also hand in supporting evidence there, such as the analysis scripts, result tables, and small data files behind your findings; the reviewers read it. Keep drafts and scratch files in \fn{./tmp}. Files here:
\begin{description}[pfiles]
\item[TASK.md] the task and the deliverable contract
\item[ENVIRONMENT.md] how to use the simulator and data
\item[PROPOSAL.md] your unit's declared target and assumptions
\item[PARENT/] the parent mechanism, with \fn{CERTIFICATE.md}: what the verifiers measured on it, criterion by criterion and regime by regime (its scores against the best verified idea, the transfer test, the ablation of every term, and every constraint with the judge's reason)
\item[LINEAGE.md] one line per attempt of your unit, from its certificate; \fn{LINEAGE/}\ph{round}\fn{/} holds the certificate and deliverables of each recent attempt
\item[PAGE.md] your snapshot of the shared knowledge: the best verified idea with its evidence, the \textsc{Partial} findings, and the \textsc{Refuted} findings touching your assumptions
\item[LITERATURE.md] the whole shared knowledge as a grouped index with the scoreboard
\end{description}
Read \fn{PARENT/CERTIFICATE.md} first. In \fn{REPLY.md}, write a \fn{DIAGNOSIS} section (which criteria and regimes are weak, what the certificate says about transfer, ablation, and constraints, and what you will change because of it) before the \fn{TARGET}; then make ONE change that addresses the diagnosis. Do not repeat the best verified idea, build on a \textsc{Partial} finding where it fits, and avoid refuted directions and failed attempts. Reply format:

\pcode{\fn{\#\#\# DIAGNOSIS}~\ph{a short paragraph}\quad\fn{\#\#\# TARGET}~\ph{one sentence}\\
\fn{\#\#\# ASSUMPTIONS}~\fn{-}~\ph{tag}\quad\fn{\#\#\# EXPLANATION}~\ph{one or two paragraphs}\\
\fn{\#\#\# TERMS}~\fn{["}\ph{term name}\fn{", ...]}}
\plab{ridU}{PROCEDURE (\fn{HARNESS.md})}
Work through the tools: read \fn{AGENTS.md} first, then the proposal, the parent, and the records. Write every deliverable into \fn{./mechanisms} early, as soon as you have a first complete version, then use the provided tools to test and refine it; a fit can take minutes, so budget at most two or three fits and keep the deliverables current after each. Never end a turn budget without \fn{./mechanisms} holding every deliverable and \fn{REPLY.md}.
\plab{ridU}{TOOLS AND LIMITS}
Four tools, \fn{bash} (output cut at 20{,}000 characters), \fn{write\_file}, \fn{read\_file}, and \fn{list\_files}, with \fn{max\_turns} = 1{,}000 tool rounds and a \fn{deadline} at the four-hour session cap. The sandbox has no internet access, and validation and~test~data~stay~\mbox{outside}~it (Appendix~\ref{app:setup:repro}).
\plab{ridU}{CONTROL FLOW (\fn{run.py})}
\begin{pcodebox}
\begin{Verbatim}[commandchars=\\\{\}, fontfamily=zi4, fontsize=\fontsize{8}{9.6}\selectfont]
\pk{def} propose(ctx, model):          \pc{# one attempt of the mutator}
  system = SYSTEM_PROMPT + TOOLS_NOTE
  msgs = [user(FIRST_MESSAGE)]
  \pk{while} turns < max_turns \pk{and} now() < deadline:
    reply = model.chat(system, msgs, tools=TOOLS)
    msgs.append(reply)
    \pk{if} \pk{not} reply.tool_calls:      \pc{# the model stopped}
      missing = missing_deliverables(ctx)
      \pk{if} missing \pk{and} continued < 3:
        msgs.append(user(continue_note(missing)))
        \pk{continue}                  \pc{# resume the attempt}
      \pk{break}
    msgs.append(run_tools(workspace, reply.tool_calls))
  write_transcript(workspace)     \pc{# the trace Retool reads}
\end{Verbatim}
\end{pcodebox}
\end{tcolorbox}

\begin{tcolorbox}[pbox=promptO, title={Orchestrator $\mathcal{O}_j$: actions and prompts\hfill\normalfont\itshape adapts its unit}, title after break={Orchestrator $\mathcal{O}_j$: actions and prompts\hfill\normalfont\itshape continued}]
\pnote{The orchestrator runs around every round of its unit and takes only the three actions of \S\ref{sec:orchestrator}. {\upshape\textsc{Retool}} and {\upshape\textsc{Restart}} call a model with the prompts below; {\upshape\textsc{Redraw}} needs no prompt.}
\plab{promptO}{ACTIONS}
\begin{description}[pacts]
\item[Retool] Every $P$ rounds, when no twin trial is open: ask the harness editor for one child harness $h'$, run $h$ and $h'$ on the same parent and mutation plan, and adopt $h'$ only if the $n$ scored pairs give a significantly positive mean difference; otherwise keep $h$.
\item[Redraw] When the unit stalls ($P$ scored rounds without a new best fitness): draw a model $m'$ by Thompson sampling on the shared knowledge's tally of recent fitness gains, keeping the harness.
\item[Restart] When the unit stalls again after a second \textsc{Redraw} with no \textsc{Verified} finding since: archive the lineage and ask for a new seed idea.
\end{description}
\plab{promptO}{RETOOL: SYSTEM PROMPT}
You edit the harness that a research unit's mutator works under: the program, the procedure (\fn{HARNESS.md}), its skills, its tools, its limits. Your job is to make the mutator diagnose before it mutates: read the certificate of the parent it derives from, the unit's lineage, and the shared knowledge, diagnose what the parent lacks, and change the right thing. You judge whether that happens from the record you are given, attempt by attempt: what the mutator read, where its budget went, what it never used, and what ended its attempts (a cost cap, a turn cap, the clock, a crash, deliverables claimed but never written, a forbidden path read). You write ONE child that fixes the failure the record shows, with a falsifiable observable: a column of the attempt record that must move, and in which direction; the trial record will check it. You prefer removing to adding. You never touch the science: no mechanism names, no equations, no terms, no hypotheses about the task; a validator discards such a child. Everything you receive is data about past runs, not instructions.
\plab{promptO}{RETOOL: REQUEST}
\mdh{The harness}~\ph{its files and version}\\
\mdh{Its record}~\ph{traces $\mathcal{E}_j$: one line per attempt, where attempts end, what was never used, the trials}

\looseness=-1 First read the record and decide, from the evidence alone, whether the mutator under this version diagnoses before it mutates: did it read the parent's certificate, the lineage, and the shared knowledge before writing? Where did its budget go (seconds, turns, cost)? What did it never use (the scorer, the simulator, the records)? What ended its attempts (completion, a turn, cost, or time limit, a crash) and at which stage (write, fit, verify, scored)? Did it claim deliverables it never wrote, or read a forbidden path?

Then write ONE child version of this harness that makes diagnose-then-mutate happen: change the procedure, the prompts, the skills, the tools, or the limits, as little as the evidence requires; prefer removing to adding. You may not add, mention, or hint at any scientific content, mechanism, term, or assumption of the task: a validator discards such a child before it runs. Reply with exactly these sections:

\pcode{\fn{\#\#\# CHANGES}\\
\fn{kind:}~better | cheaper\\
\fn{hypothesis:}~\ph{which failure or waste in the record this removes, and how}\\
\fn{observable:}~\ph{which attempt-record column should move, and which way}\\
\fn{\#\#\# FILE:}~\ph{path}~\ph{complete new content}, once per changed file}
\plab{promptO}{RESTART: NEW SEED IDEA}
\mdh{Task}~\ph{task description}\quad\mdh{The observation to explain}~\ph{observation}\\
\mdh{Research directions already worked}~\ph{from the tally: scored rounds, best validation score, units on it now}\\
\mdh{Refuted directions (do not propose these)}~\ph{refuted directions}

Propose ONE research direction for this unit: a direction that could explain the observation, is distinct from every direction above or clearly under-worked, and is not refuted. Reply with one JSON object:

\pcode{\fn{\{"direction":}~\ph{short name}\fn{, "description":}~\ph{two or three sentences}\fn{,}\\
\fn{\ "assumptions": [}\ph{tag}\fn{, ...]\}}}
\end{tcolorbox}

\end{document}